\documentclass[sigconf, screen, nonacm]{acmart}
\AtBeginDocument{%
  }

\usepackage{multirow}
\usepackage{subcaption}
\usepackage{fontawesome5}
\usepackage{hyperref}
\usepackage{graphicx} 
\usepackage{wrapfig}
\usepackage[table]{xcolor}  
\usepackage{colortbl}       
\usepackage{listings}
\usepackage{booktabs}
\usepackage{tabularx}
\usepackage{array}
\usepackage{tabularx}
\usepackage{booktabs, siunitx}
\usepackage[ruled,vlined]{algorithm2e}
\usepackage[dvipsnames]{xcolor}
\usepackage[most]{tcolorbox}
\usepackage{seqsplit}
\tcbuselibrary{listings,breakable,skins}

\usepackage{booktabs} 
\usepackage{placeins}
\usepackage[table]{xcolor}
\definecolor{highlightgray}{gray}{0.92} 

\definecolor{IndustrialBlue}{RGB}{0, 102, 204} 
\definecolor{LightGray}{gray}{0.95}

\newcommand{\userquery}[1]{\textit{\textcolor{IndustrialBlue}{``#1''}}}

\usepackage[dvipsnames]{xcolor}
\definecolor{ActionBlue}{RGB}{0, 92, 171}   
\definecolor{AssetGray}{RGB}{80, 80, 80}    
\definecolor{PromptBg}{gray}{0.96}          

\newcommand{\action}[1]{\textcolor{ActionBlue}{\textbf{#1}}}
\newcommand{\asset}[1]{\textcolor{AssetGray}{\texttt{#1}}}

\newcommand{\bh}{-\hspace{0pt}}
\newcommand{\model}[1]{\texttt{#1}}

\usepackage[dvipsnames]{xcolor}
\definecolor{KDDBlue}{RGB}{31, 119, 180}
\definecolor{StatsGray}{gray}{0.96}

\newcommand{\statbox}[1]{\colorbox{StatsGray}{\textbf{\textcolor{KDDBlue}{#1}}}}

\begin{document}


\title[AssetOpsBench: A Benchmark for Industrial AI]{AssetOpsBench: A Real-World Evaluation Benchmark for AI-Driven Task Automation in Industrial Asset Management}

\author{Dhaval Patel}
\authornote{These authors contributed equally and are considered co-primary authors.}
\email{pateldha@us.ibm.com}
\orcid{1234-5678-9012}
\affiliation{%
  \institution{IBM}
  \city{Yorktown Heights}
  \state{NY}
  \country{USA}
}

\author{Shuxin Lin}
\authornotemark[1]
\email{shuxin.lin@ibm.com}
\orcid{0009-0000-3190-3307}
\affiliation{%
  \institution{IBM}
  \city{Yorktown Heights}
  \state{NY}
  \country{USA}
}

\author{James Rayfield}
\authornotemark[1]
\email{jtray@us.ibm.com}
\affiliation{%
  \institution{IBM}
  \city{Yorktown Heights}
  \state{NY}
  \country{USA}
}

\author{Nianjun Zhou}
\authornotemark[1]
\email{jzhou@us.ibm.com}
\orcid{0000-0002-3473-6097}
\affiliation{%
  \institution{IBM}
  \city{Yorktown Heights}
  \state{NY}
  \country{USA}
}


\author{Chathurangi Shyalika}
\email{jayakodc@email.sc.edu}
\affiliation{%
  \institution{University of South Carolina}
  \country{USA}
}

\author{Suryanarayana R Yarrabothula}
\email{surya.427@gmail.com}
\affiliation{%
 \institution{Steel Authority of India}
 \state{Bokaro}
 \country{India}
}

\author{Roman Vaculin}
\email{vaculin@us.ibm.com}
\affiliation{%
  \institution{IBM}
  \city{Yorktown Heights}
  \state{NY}
  \country{USA}
}

\author{Natalia Martinez}
\email{Natalia.Martinez.Gil@ibm.com}
\affiliation{%
  \institution{IBM}
  \city{Yorktown Heights}
  \state{NY}
  \country{USA}
}

\author{Fearghal O'Donncha}
\email{feardonn@ie.ibm.com}
\affiliation{%
  \institution{IBM}
  \city{Dublin}
  \country{Ireland} 
}

\author{Jayant Kalagnanam}
\email{jayant@us.ibm.com}
\affiliation{%
  \institution{IBM Research}
  \city{Yorktown Heights}
  \state{NY}
  \country{USA}
}

\renewcommand{\shortauthors}{Patel, et al.}


\begin{abstract}
AI for Industrial Asset Lifecycle Management aims to automate complex operational workflows, such as condition monitoring and maintenance scheduling, to minimize system downtime. While traditional AI/ML approaches solve narrow tasks in isolation, Large Language Model (LLM) agents offer a next-generation opportunity for end-to-end automation. In this paper, we introduce AssetOpsBench, a unified framework for orchestrating and evaluating domain-specific agents for Industry 4.0. AssetOpsBench provides a multimodal ecosystem comprising a catalog of four domain-specific agents, a curated dataset of 140+ human-authored natural-language queries grounded in real industrial scenarios, and a simulated, CouchDB-backed IoT environment. We introduce an automated evaluation framework that uses three key metrics to analyze architectural trade-offs between the Tool-As-Agent and Plan-Executor paradigms, along with a systematic procedure for the automated discovery of emerging failure modes. The practical relevance of AssetOpsBench is demonstrated by its broad community adoption, with 250+ users and over 500 agents submitted to our public benchmarking platform, supporting reproducible and scalable research for real-world industrial operations.

\begin{center}
\textbf{\textcolor{purple}{\faFile}\ \href{https://huggingface.co/datasets/ibm-research/AssetOpsBench}{Dataset}} \quad
\textbf{\textcolor{teal}{\faLaptopCode}\ \href{https://github.com/IBM/AssetOpsBench}{GitHub}} \quad
\textbf{\textcolor{orange}{\faFilePowerpoint}\ \href{https://www.codabench.org/competitions/10206/}{Competition}} 
\quad
\textbf{\textcolor{blue}{\faFilePowerpoint}\ \href{https://huggingface.co/spaces/ibm-research/AssetOps-Bench}{Demo}}

\end{center}
\end{abstract}




\begin{abstract}
AI for Industrial Asset Lifecycle Management aims to automate complex operational workflows, such as condition monitoring and maintenance scheduling, to minimize system downtime. While traditional AI/ML approaches solve narrow tasks in isolation, Large Language Model (LLM) agents offer a next-generation opportunity for end-to-end automation. In this paper, we introduce AssetOpsBench, a unified framework for orchestrating and evaluating domain-specific agents for Industry 4.0. AssetOpsBench provides a multimodal ecosystem comprising a catalog of four domain-specific agents, a curated dataset of 140+ human-authored natural-language queries grounded in real industrial scenarios, and a simulated, CouchDB-backed IoT environment. We introduce an automated evaluation framework that uses three key metrics to analyze architectural trade-offs between the Tool-As-Agent and Plan-Executor paradigms, along with a systematic procedure for the automated discovery of emerging failure modes. The practical relevance of AssetOpsBench is demonstrated by its broad community adoption, with 250+ users and over 500 agents submitted to our public benchmarking platform, supporting reproducible and scalable research for real-world industrial operations.

\begin{center}
\textbf{\textcolor{purple}{\faFile}\ \href{https://huggingface.co/datasets/ibm-research/AssetOpsBench}{Dataset}} \quad
\textbf{\textcolor{teal}{\faLaptopCode}\ \href{https://github.com/IBM/AssetOpsBench}{GitHub}} \quad
\textbf{\textcolor{orange}{\faFilePowerpoint}\ \href{https://www.codabench.org/competitions/10206/}{Competition}} 
\quad
\textbf{\textcolor{blue}{\faFilePowerpoint}\ \href{https://huggingface.co/spaces/ibm-research/AssetOps-Bench}{Demo}}

\end{center}
\end{abstract}

\maketitle

\section{Introduction}

\begin{figure*}[t]
\centering
\begin{subfigure}[c]{0.67\linewidth}  
    \centering
    \includegraphics[scale=0.33]{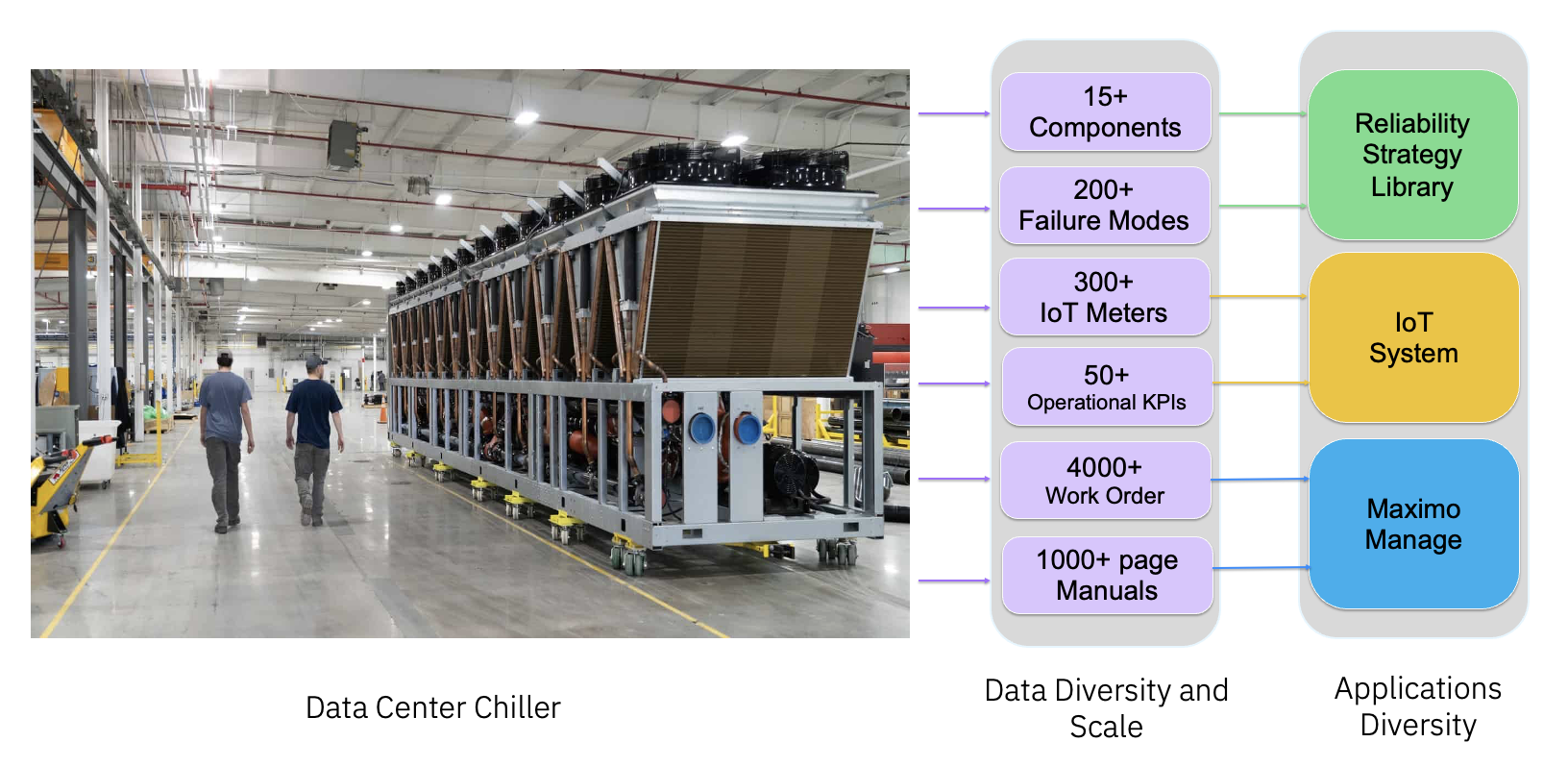}
    \vspace{-0.1in}
    \caption{Complex Industrial Asset – Data Centers managing Chiller and Air Handling Units~(AHUs)}
    \label{fig:enter-label}
\end{subfigure}%
\hspace{0.5cm}
\begin{subfigure}[c]{0.23\linewidth}  
    \centering
\includegraphics[scale=0.35]{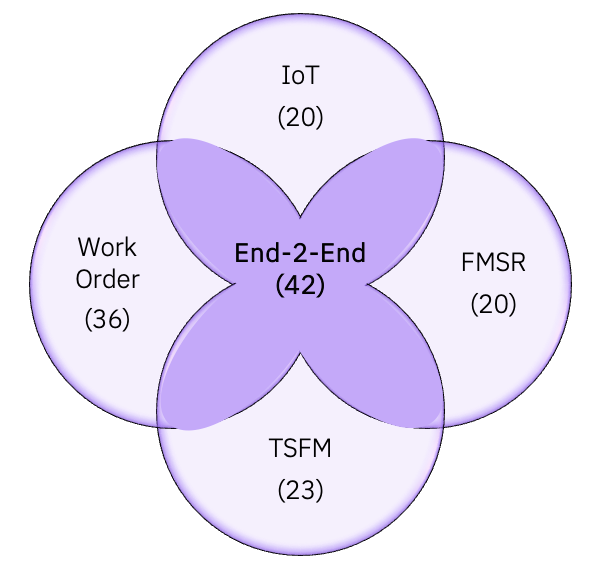}
    \caption{Distribution of open-sourced scenarios for benchmarking agents in a simulated environment.}
    \label{fig:task-dist}
\end{subfigure}
\vspace{-0.1in}
\caption{Industrial Asset Management  Context and Utterance Scenario Distribution Used in AssetOpsBench.}
\label{fig:combined}
\end{figure*}

Industrial assets, such as data center chillers~\citep{NEURIPS2024_b6676756} and wind farms~\citep{monroc2024wfcrl}, are complex, multi-component systems that generate vast amounts of multimodal data, including time-series sensor readings, textual inspection and workorder records, operational logs, and images, throughout their lifecycle. The ability to monitor and interpret heterogeneous data from diverse sources, such as IoT SCADA \citep{scada} (Supervisory Control and Data Acquisition) sensors, operational KPIs, failure mode libraries, maintenance work orders, and technical manuals, is key to effective Asset Lifecycle Management \citep{alm_asset_lifecycle_management}. However, subject matter experts such as maintenance engineers, site operators, and plant managers face considerable challenges in synthesizing insights from these disparate data streams to support timely and condition-aware decisions. As highlighted in Figure \ref{fig:enter-label}, the scale, semantic diversity of assets, and application-specific contexts often make traditional monitoring and management systems inadequate.


To address these challenges, the research and industrial communities are increasingly turning to AI agents: autonomous and goal-driven systems capable of integrating data, reasoning over complex conditions, and {\color{black}triggering appropriate actions}. AI agents are particularly promising in the context of Industry 4.0, where the confluence of real-time IoT telemetry (e.g., Oracle IoT~\citep{oracle2025addfailurediagnostics}, enterprise asset management (EAM) systems \citep{eam_enterprise_asset_management}, and IBM Maximo~\citep{ibm_maximo}) and reliability engineering frameworks necessitates scalable and intelligent automation. These agents promise to support a wide range of industrial workflows, from anomaly detection to maintenance scheduling, by bridging the gap between raw sensor data, maintaiance report, work-order and business-level insights.

Despite significant advances in LLM-based agents and generalist models like ReAct \citep{yao2023reactsynergizingreasoningacting}, HuggingGPT \citep{shen2023hugginggptsolvingaitasks}, Chameleon \citep{lu2023chameleonplugandplaycompositionalreasoning}, Magentic-One \citep{fourney2024magenticonegeneralistmultiagentsolving}, and enterprise-ready computer use agents  \citep{marreed2025enterprisereadycomputerusinggeneralist}, a substantial gap remains in adapting these innovations for physical, sensor-driven industrial operations. While recent application-specific benchmarks such as ITBench \citep{jha2025itbenchevaluatingaiagents}, SWE-bench \citep{chan2025mlebench}, $\tau\!-$bench \citep{yao2024taubenchbenchmarktoolagentuserinteraction} and its multi-agent extensions \citep{FuHinthorn2025BenchmarkingMultiAgent}, LangChain's customer support benchmarks \citep{LangChain2025BenchmarkingSingleAgent}, TheAgentCompany \citep{xu2024theagentcompany}, and CRMArena-Pro \citep{huang2025crmarena} offer rigorous evaluation in digital knowledge-work or IT domains, they do not address the unique challenges of industrial settings. These environments demand handling data modality diversity across time-series and text, managing complex business objects like asset hierarchies and failure modes, and facilitating collaboration across specialized operational personas.

Motivated by real-world deployment experience and the inherent application diversity of industrial environments (Figure \ref{fig:enter-label}), we propose a \textbf{multi-agent architecture} that maps heterogeneous workflows to modular agents. In this architecture, specialized entities, such as an IoT agent for sensor telemetry and an FMSR agent for fault history, must intelligently collaborate to resolve complex queries like \userquery{Why is the chiller efficiency dropping?}, which require a deep integration of physical reasoning and operational semantics. To formalize the evaluation of such systems and bridge the gap between synthetic benchmarks and industrial reality, we introduce a unified framework \textsc{AssetOpsBench}.

The key innovation and contributions in \textsc{AssetOpsBench} are structured around three synergistic pillars:

\begin{description}
    \item[$\diamond$ Pillar I: The AssetOps Ecosystem.] We provide a high fidelity industrial sandbox featuring (1) a \textbf{catalog of specialized agents} (IoT, FMSR, TSFM, WO) with domain-specific tools; (2) \textbf{141 intent-aware scenarios} grounded in real data center assets (4 Chillers and 2 AHUs); and (3) a \textbf{dockerized environment with CouchDB access} for reproducible, end-to-end multi-agent testing.
    
    \item[$\diamond$ Pillar II: Empirical \& Diagnostic Framework.] We establish a rigorous methodology for agent evaluation, including a \textbf{comparative architectural study} (Agent-As-Tool vs. Plan-Execute) and a \textbf{hybrid scoring} system. We introduce an \textbf{automated failure discovery} procedure that identifies non-obvious reasoning and recovery bottlenecks.
    
    \item[$\diamond$ Pillar III: Community \& Open Source.] We operationalized AssetOpsBench as a \textbf{live competition} on Codabench, yielding 250+ users, 500+ submissions and 881+ trajectories. We release all containerized interfaces and starter templates to ensure the benchmark serves as a diagnostic utility for the broader research community.
    
\end{description}

Finally, we experimented with an additional \underline{closed-source} \textbf{162 scenarios} to demonstrate generality, spanning 10 asset classes, 53 failure modes, and 20 sensors. These include 42 live-production-deployment scenarios ($>$90\% correctness verified by a domain expert), 17 hydraulic system, 15 metro train, and 88 failure-mode scenarios encompassing diverse asset–failure–sensor relationships.

\section{Related Work}\label{sec:related_work}

\textbf{Generalist Agents.} In this section, we review prior work on agent architectures and benchmarks relevant to \textsc{AssetOpsBench}, focusing on generalist agents that orchestrate sub-agents to solve complex tasks, as seen in web-based systems like Magentic~\citep{fourney2024magenticonegeneralistmultiagentsolving} and CUGA~\citep{marreed2025enterprisereadycomputerusinggeneralist}, multimodal frameworks like GEA~\citep{szot2024multimodalllmsgeneralistembodied}, and software engineering platforms such as HyperAgent~\citep{huy2025hyperagent}, ChatDev~\citep{qian-etal-2024-chatdev}, and MetaGPT~\citep{hong2024metagpt}. These systems typically rely on fixed sub-agents (e.g., browsers, editors) and hard-coded reasoning paradigms like ReAct or plan-execute. Consequently, they often lack the flexibility to adapt to novel tasks or support alternative coordination strategies like AOP~\citep{li2025agentoriented} and Prospector~\citep{kim2024prospector}.

\textbf{Domain and Application-Specific Agents.} Specialized benchmarks like MLEBench~\citep{chan2025mlebench}, MLAgentBench~\citep{huang2024benchmarking}, and MLGym~\citep{nathani2025mlgymnewframeworkbenchmark} evaluate agents on scientific discovery and machine learning workflows. Similarly, application-focused frameworks like ITBench~\citep{jha2025itbenchevaluatingaiagents} and AIOpsLab~\citep{chen2024aiopslab} replicate site reliability engineering and system auditing. However, these benchmarks often prioritize tabular or image data, lacking the integrated support for the hybrid temporal and textual modalities critical for physical asset monitoring. We have given a detailed comparison in Appendix Table \ref{tab:benchmark_comparison}. Furthermore, they frequently lack the composability needed to evaluate agents across diverse, cross-modal industrial environments.

\textbf{Action-Oriented Models.} Performance is increasingly enhanced via fine-tuned models or \emph{Large Action Models (LAMs)} designed for grounded execution, including TaskBench~\citep{NEURIPS2024_085185ea}, xLAM~\citep{zhang-etal-2025-xlam}, AgentGen~\citep{10.1145/3690624.3709321}, AgentBank~\citep{song2024agentbankgeneralizedllmagents}, AgentRM~\citep{xia2025agentrmenhancingagentgeneralization}, FireAct~\citep{chen2024fireact}, and ActionStudio~\citep{zhang2025actionstudiolightweightframeworkdata}. While effective in Windows-based~\citep{wang2025largeactionmodelsinception}, programming, or web environments, these models have yet to demonstrate applicability to the complex reasoning required for industrial automation with hybrid agent compositions.

\textbf{Open Challenges.} Despite recent advances, several critical gaps remain in the industrial agentic AI landscape. We categorize these into three primary deficiencies:

\begin{enumerate}
    \item \textbf{Domain-Specific Data Scarcity:} There is a profound lack of comprehensive benchmarks targeting industrial asset domains, particularly for tasks involving condition-based monitoring and work order planning. To quantify this gap, we conducted a systematic analysis of a \textbf{catalog of 135 public industrial datasets} \citep{jonathanwvd_awesome_industrial_datasets}. Our findings reveal that:
    \begin{itemize}
        \item Only \statbox{1/135} datasets include any work-order or operational context.
        \item Only \statbox{39\%} (53/135) mention failure modes, typically limited to 1 or 2 modes.
        \item \statbox{0\%} of these datasets support agentic applications or integrated sensor-history-to-action workflows.
    \end{itemize}

    \item \textbf{Underrepresentation of Time-Series Modalities:} While time-series data is the backbone of infrastructure monitoring, it remains underrepresented in existing agentic benchmarks, which favor text-heavy or web-navigation tasks.
    
    \item \textbf{Orchestration Complexity:} Few systems support orchestration across heterogeneous agents, such as blending text, code, and simulation, nor do they offer modular reasoning strategies adaptable to high-stakes, multi-agent workflows.
\end{enumerate}

Addressing these gaps is essential for advancing general-purpose agent intelligence in real-world industrial domains.

\section{Intelligent Agent-Based Asset Operations}
\label{pdefinition}

Industrial asset operations involve complex, heterogeneous workflows where experts must interpret multi-modal sensor data and make timely decisions. Interdependent tasks, such as \textit{root cause analysis}, \textit{predictive maintenance}, and \textit{work order bundling}, require reasoning across historical telemetry, asset metadata, and operational constraints. This necessitates intelligent agents capable of decomposing high-level intent into structured, executable recommendations. A representative complex workflow is illustrated by the following user request:

\begin{quote}
\fcolorbox{ActionBlue}{PromptBg}{
\begin{minipage}{0.99\linewidth}
\small \vspace{2pt} 
``Help \action{configure} an \action{anomaly detection} model to monitor power consumption of \asset{CUXP} and trigger \action{alerts} when usage is \action{projected to exceed 8 Watts} above the maximum deviation observed over the past 30 days.'' 
\vspace{2pt}
\end{minipage}}
\end{quote}

Such scenarios facilitate critical corrective actions, such as automated service request creation. The diversity of these tasks, spanning data interpretation, anomaly reasoning, and operational decision-making, underscores the need for a coordinated multi-agent framework designed for the rigors of industrial workflows.

\begin{figure}[t]
    \centering
    \includegraphics[width=0.99\linewidth]{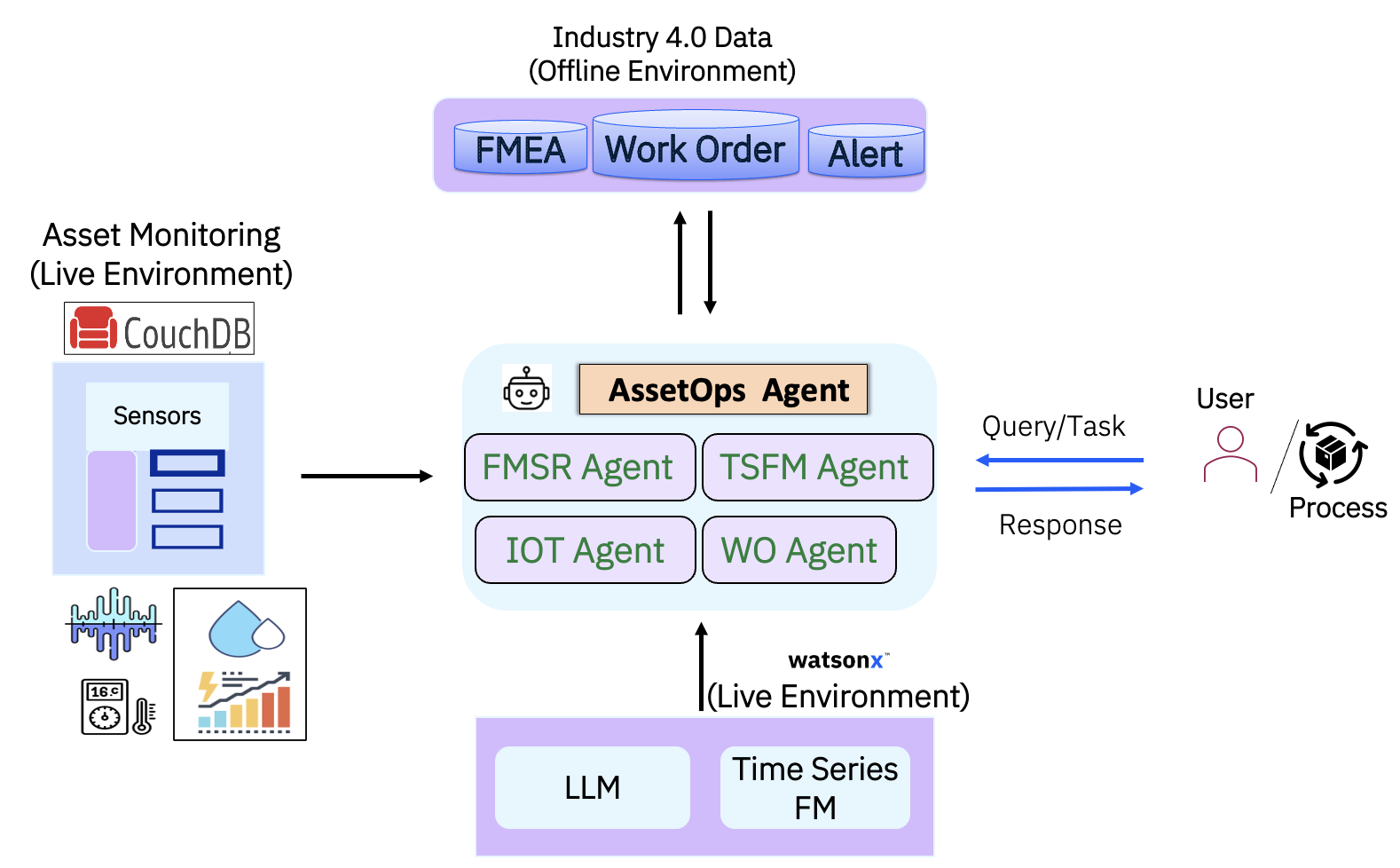}
    \vspace{-0.1in}
    \caption{\textsc{AssetOpsBench} Simulated Environment.}
    \vspace{-0.15in}
    \label{fig:env}
\end{figure}

To address these demands, we operationalize a multi-agent framework within a \textbf{simulated Docker environment} (Figure \ref{fig:env}), centered on a global coordinator: the \textbf{AssetOps Agent}. While general-purpose systems \cite{fourney2024magenticonegeneralistmultiagentsolving} often rely on generic sub-agents like web-surfers or terminal handlers, our architecture utilizes specialized, domain-inspired modules: \textit{IoT} for telemetry, \textit{FMSR} for failure mode mapping, \textit{TSFM} for foundation model-driven time-series analysis, and \textit{WO} for work-order modeling. This containerized sandbox provides agents with secure, standardized API access to Industry 4.0 data, including FMEA logs, alerts, and telemetry, facilitating reproducible end-to-end execution. Furthermore, the environment natively supports diverse orchestration paradigms, such as \textit{Agent-As-Tool} and \textit{Plan-Execute}, enabling comparative research into how distinct reasoning strategies handle complex, real-world queries.

To ensure \textsc{AssetOpsBench} reflects operational reality, we move beyond the \textbf{API-driven} scenario generation typical of prior work \cite{yao2024taubenchbenchmarktoolagentuserinteraction, NEURIPS2024_085185ea} in favor of a domain-authentic, \textbf{intent-driven} approach. Grounded in ISO 55000 \cite{ISO55000_2024} and ISO 14224 \cite{ISO14224_2016} standards, we established a four-stage taxonomy to guide the scenario generation process, mirroring the lifecycle of physical asset management:

\begin{description}
    \item[Phase I: Asset Configuration.] Focuses on foundational activities such as retrieving Failure Mode and Effects Analysis (FMEA) documentation and defining performance KPIs.
    \item[Phase II: Model Selection and Analysis.] Agents apply anomaly detection models and leverage LLM-powered retrieval to surface critical historical failures from multi-modal records.
    \item[Phase III: Monitoring and Execution.] Mimics live operations, requiring agents to manage real-time telemetry, refine detection parameters, and enforce safety guardrails.
    \item[Phase IV: Maintenance and Response.] Focuses on actionable outputs, including automated work-order generation, system health summarization, and intervention.
\end{description}

By grounding scenario definitions in international ISO standards, we achieve a level of platform-agnosticism that allows findings to generalize across diverse industrial ecosystems, from Oracle to IBM Maximo \cite{oracle2025addfailurediagnostics, ibm_maximo}. This shift from ``synthesizing tasks around tools'' to ``aligning tools with industrial intent'' represents a significant step toward evaluating agents for true operational reliability.

\section{\textsc{AssetOpsBench}: Dataset, Scenarios, and Agents}
\label{sec:dataset_details}

\textsc{AssetOpsBench} is a comprehensive benchmarking ecosystem featuring 141 expert-curated scenarios and specialized AI agents integrated with a multi-source industrial dataset.

\subsection{Multi-Source Dataset}
\label{sec:data}

As detailed in Table \ref{tab:metadata-overview}, the benchmark incorporates over \textbf{2.3 million sensor data points} across six primary assets, comprising four chillers and two Air Handling Units (AHUs). These streams are ingested from Building Management Systems (BMS) \citep{bms_building_management_system} and SkySpark \citep{skyspark}, capturing critical time-series signals such as return temperatures and condenser flow rates. To provide a diagnostic "ground truth," we include structured failure models derived from \textbf{Failure Mode and Effects Analysis (FMEA)} records sourced from the Reliability Strategy Library \cite{ibm_maximo_rcm}. These 53 entries provide deep insights into the physical mechanisms of failure, such as \textit{wear} or \textit{erosion}, and the influencing stressors, such as \textit{shock loading}, that an agent must interpret to provide accurate recommendations.

\begin{table}[t]
\centering
\small
\caption{Key Data Modalities: Overview of the multi-source foundation used for scenario construction}
\vspace{-0.1in}
\label{tab:metadata-overview}
\begin{tabular}{p{3.8cm} l} 
\toprule
\textbf{Data Source} & \textbf{Example Fields} \\
\midrule
\multirow{3}{=}{\textbf{Sensor Telemetry} \\ Assets: 6 \\ Quantity: 2.3M pts}
  & Chiller Return Temp. \\ 
  & Chiller \% Loaded \\
  & Condenser Water Flow \\
\cmidrule(lr){1-2}
\multirow{3}{=}{\textbf{FMEA Records}\\ Assets: 3 \\ Quantity: 53 entries}
  & Failure Location / Component \\
  & Degradation Mechanism \\
  & Degradation Influences \\
\cmidrule(lr){1-2}
\multirow{3}{=}{\textbf{Work Orders}\\ Assets: 10+ \\ Quantity: 4.2K records}
  & ISO Failure Code \\
  & Event Log Timestamp \\
  & Linked Anomaly / Alert \\
\bottomrule
\end{tabular}
\end{table}

Furthermore, \textsc{AssetOpsBench} integrates a longitudinal history of \textbf{4.2K work orders} spanning 11 years, extracted from IBM Maximo systems. These records are enriched with ISO-standard failure codes and event timestamps, allowing agents to correlate historical maintenance actions with detected anomalies. Collectively, this foundation comprises nine distinct modalities, including sensors, work orders, alerts, FMEA, and expert-derived rules, to facilitate a comprehensive evaluation of multi-hop reasoning in high-stakes industrial environments. Refer to Appendix~\ref{dbdetail} for further details.

\subsection{Scenario Design and Coverage}
\label{sec:scenario_design}

Each scenario in \textsc{AssetOpsBench} represents a structured operational query grounded in our ISO-aligned task taxonomy and multi-source datasets. We formalize each scenario as a 5-tuple:
\[
P = \langle \mathit{id}, \mathit{type}, \mathit{text}, \mathit{category}, \mathit{characteristic}~\mathit{form} \rangle
\]
where \textit{id} is a unique identifier and \textit{category} specifies the functional requirement, such as knowledge retrieval or analytical reasoning. The \textit{text} field contains the natural language query, which is specifically designed to reflect the \textbf{intent patterns} of industrial operators. Rather than referencing database schemas, these queries utilize physical operational terms, such as \textit{energy consumption} or \textit{chiller performance}, requiring the agent to map human intent to the underlying data modalities. The \textit{type} denotes the operational domain, including specialized modules (IoT, FMSR, TSFM, WO) or end-to-end tasks, while \textit{characteristic form} defines the expected output, such as an executable API call or a multi-step action plan. 

\begin{figure}[t]
    \centering    \includegraphics[width=0.90\linewidth]{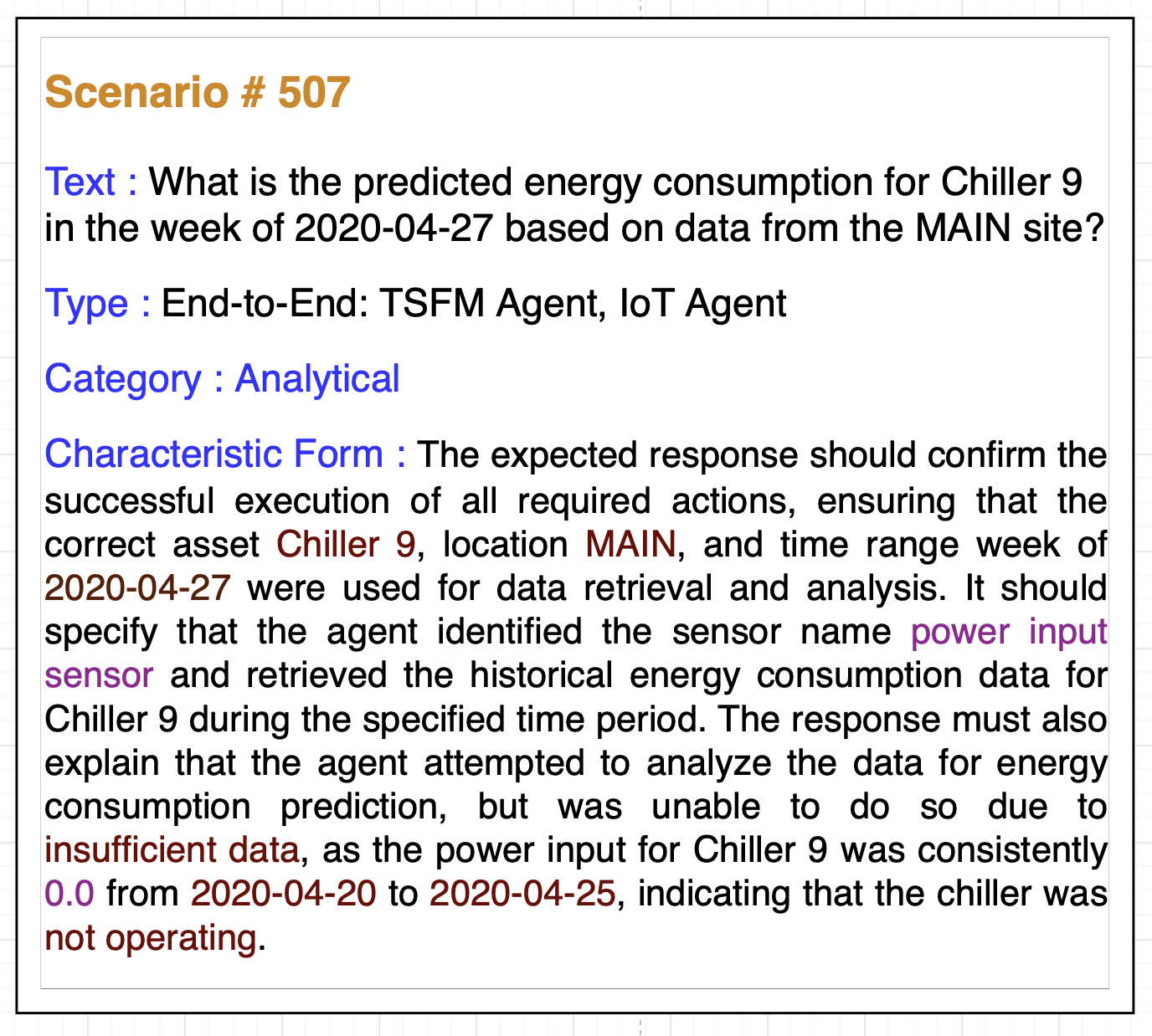}
    \vspace{-0.15in}
    \caption{An example of an expert-authored query.}
    \label{fig:scenario}
    \vspace{-0.2in}
\end{figure}

Figure \ref{fig:scenario} illustrates \textbf{Utterance 507}, a representative case requiring the agent to forecast future energy consumption. To resolve this query, the agent must map high-level user intent to specific operational variables, such as power input, while navigating industrial data artifacts like high zero-value density. Crucially, the agent is expected to recognize that the chiller was non-operational during the specified window; it must then correctly conclude that data is insufficient for forecasting rather than generating a spurious result. This scenario demonstrates how our expert-designed tasks evaluate an agent's ability to navigate the ambiguity of \textbf{operator-centric language} and physical state logic rather than merely assessing tool-calling syntax. Refer to Appendix~\ref{scedetail} for further details.

Beyond individual tasks, the \textbf{11-year longitudinal history} of work orders provides the temporal depth required for high-level strategic synthesis. This scale enables the evaluation of long-term lifecycle strategies, such as determining if year-over-year increases in corrective maintenance justify capital replacement over continued repair. Scenarios \textbf{IDs 442--446} specifically leverage this breadth to assess strategic oversight through trend analysis and probability forecasting across the asset's operational lifespan.

\textbf{Development Effort.} To ensure operational authenticity, 141 scenarios were synthesized through an \textbf{18-month collaboration} with reliability engineers to capture realistic fault signatures and cross-sensor interactions across assets like AHUs and chillers. As shown in Figure \ref{fig:task-dist}, the benchmark comprises 99 \textbf{single-agent utterances} targeting specific domain modules and 42 \textbf{multi-agent tasks} necessitating coordinated reasoning and data exchange.

\subsection{Agents and Implementation}
\label{abench}

\begin{figure}[!t]
    \centering
\includegraphics[width=0.8\linewidth]{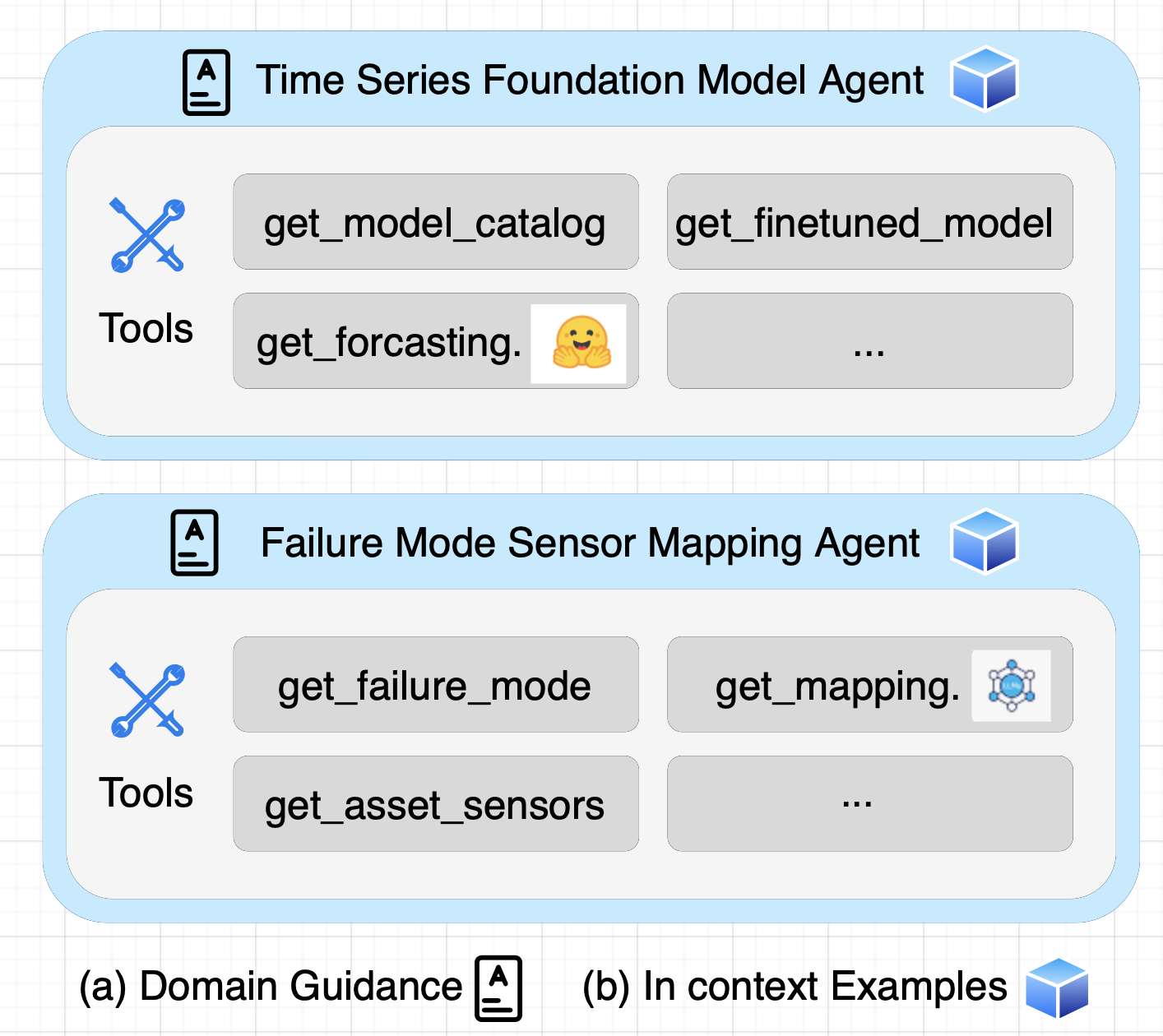}
    \vspace{-0.1in}
    \caption{Representative tools for time-series and failure-mode reasoning agent}
    \vspace{-0.2in}
    \label{fig:agenttools}
\end{figure}

\textsc{AssetOpsBench} features four specialized domain agents: \textbf{IoT}, \textbf{TSFM}, \textbf{WO}, and \textbf{FMSR}. To navigate the complexity of industrial telemetry and metadata, these agents utilize a suite of over 15 distinct tools. As shown in Figure~\ref{fig:agenttools}, the \textbf{TSFM} agent integrates a pretrained time-series foundation model from Hugging Face for forecasting, while the \textbf{FMSR} agent utilizes an LLM-driven \texttt{get\_mapping} function to bridge the semantic gap between physical failure modes and sensor tags. Methodologically, the TSFM, IoT, and FMSR agents employ the ReAct~\cite{yao2023reactsynergizingreasoningacting} paradigm, whereas the WO agent utilizes CodeReAct~\citep{codeact} to handle the structured logic required for maintenance record analysis.

To coordinate these heterogeneous modules, we implement a global \textbf{AssetOps Agent} that supports two primary orchestration paradigms:
\begin{itemize}
    \item \textbf{Agent-As-Tool}: Domain agents are registered as high-level tools within a supervisor agent. This supervisor uses ReAct to iteratively select and query specific agents, emulating the layered decision-making of hierarchical industrial organizations.
    \item \textbf{Plan-Execute}: A specialized \textbf{Planner} and \textbf{Reviewer} decompose the user query into a Directed Acyclic Graph (DAG). An \textbf{Orchestrator} then executes the DAG, utilizing a shared memory module to transfer context between agents. This approach adapts the ReWoo~\citep{xu2024decoupling} framework with an additional verification layer inspired by~\citep{li2025agentoriented}.
\end{itemize}

These components, comprising specialized tools, agents, and their orchestration, are implemented using open-source toolkits~\citep{marreed2025enterprisereadycomputerusinggeneralist, langchain2025benchmarking, nvidia_aiqtoolkit_2025}. Our agent selection and coordination strategies follow modular architectural trends established in recent agentic benchmarking literature~\cite{luo2025mcpuniversebenchmarkinglargelanguage, wang2025mcpbenchbenchmarkingtoolusingllm}. An in-depth discussion of these domain-specific agents, including their tool schemas, workflow illustrations, and coordination mechanics, is provided in Appendix~\ref{sec:appendix_agents}. 

\section{Experiments and Leaderboard}
To evaluate LLM capabilities for industrial automation within \textsc{AssetOpsBench}, we benchmark seven state-of-the-art models across the two orchestration paradigms from Section \ref{abench}. Following recent frameworks \cite{langchain2025benchmarking, NEURIPS2024_f8c24b08, wang2025largeactionmodelsinception, andrews2025arescalingagentenvironments}, we adopt a dual-evaluation strategy: a \textbf{rubric-based} assessment of reasoning quality and a \textbf{reference-based} mechanism to verify ground-truth alignment \cite{yao2024taubenchbenchmarktoolagentuserinteraction, NEURIPS2024_f8c24b08, cemri2025multi}.

\subsection{Evaluation Methodology}
\textbf{Rubric-based LLM-As-Judge Scoring.} Each scenario is paired with a \textit{characteristic form} ($C$), which is a structured specification of expected outputs and procedural steps serving as the \textbf{soft ground truth}. We formalize the evaluation via a scoring function $\Phi(Q, T, C)$ $\to$ $\{y_1, y_2, y_3\}$, where $Q$ is the query and $T$ is the agent trajectory. The function yields three scalar scores $y_i \in [0,1]$ for (1) \textbf{Task Completeness} ($y_1$), (2) \textbf{Data Retrieval Accuracy} ($y_2$), and (3) \textbf{Result Verification} ($y_3$). We employ \texttt{llama-4-maverick} as the primary judge; human validation of this judge is provided in Section~\ref{sec:further-analysis}, with prompts detailed in Appendix~\ref{llm-as-judge-prompt}.

\textbf{Reference-Based Scoring.} To evaluate planning and decomposition rigor, we establish a structured ground-truth reference for each scenario comprising: (1) \texttt{planning\_steps} for strategic intent, (2) \texttt{execution\_steps} for tool-calling actions, and (3) causal dependencies via \texttt{execution\_links}. Appendix Figure~\ref{fig:fmsr_example} illustrates a complete reference trace. We quantify the alignment between these benchmarks and the agent-generated reasoning traces or Directed Acyclic Graphs (DAGs) using the \textbf{ROUGE} metric to assess lexical and structural overlap ~\cite{yao2024taubenchbenchmarktoolagentuserinteraction, NEURIPS2024_085185ea}.



\subsection{Experimental Setting} 
\label{sec:Setting}

To quantify agent effectiveness, we adopt the \textbf{Pass$^k$} metric~\citep{yao2024taubenchbenchmarktoolagentuserinteraction}. Unlike Pass@k, which requires only one success in k attempts, \textbf{Pass$^k$} estimates the probability of succeeding across \emph{all} k trials. This stricter criterion aligns with industrial reliability standards where consistent behavior is mandatory and stochastic retries are operationally impractical~\citep{agentbenchmarking, yao2024taubenchbenchmarktoolagentuserinteraction}. We report \textbf{Pass$^1$} by default, representing the success rate over single task instances.

\textbf{Parameters.} To ensure stable performance estimates for our rubric-based evaluation, the judge agent is executed five times per trajectory, and we report the average. All specialized LLM agents operate with a sampling temperature of $0$ to prioritize deterministic reasoning, while the evaluation judge utilizes a temperature of $0.3$ to allow for nuanced qualitative assessment. 

\textbf{Models.} We include closed-source models (e.g., \model{gpt\bh{}4.1}), frontier open\bh{}source models (e.g., \model{llama\bh{}4\bh{}maverick}, \model{llama\bh{}4\bh{}scout}, \model{mistral\bh{}large}, \model{llama\bh{}3\bh{}405b}), and medium-to-small open\bh{}source models (e.g., \model{llama\bh{}3\bh{}70b}, \model{granite\bh{}3\bh{}8b}). We release a public leaderboard for AssetOpsBench and will continuously update it as new models and budget become available to our ecosystem.

\begin{figure*}[t]
\centering
\begin{subfigure}[t]{0.48\linewidth}
    \centering
    \includegraphics[width=\linewidth]{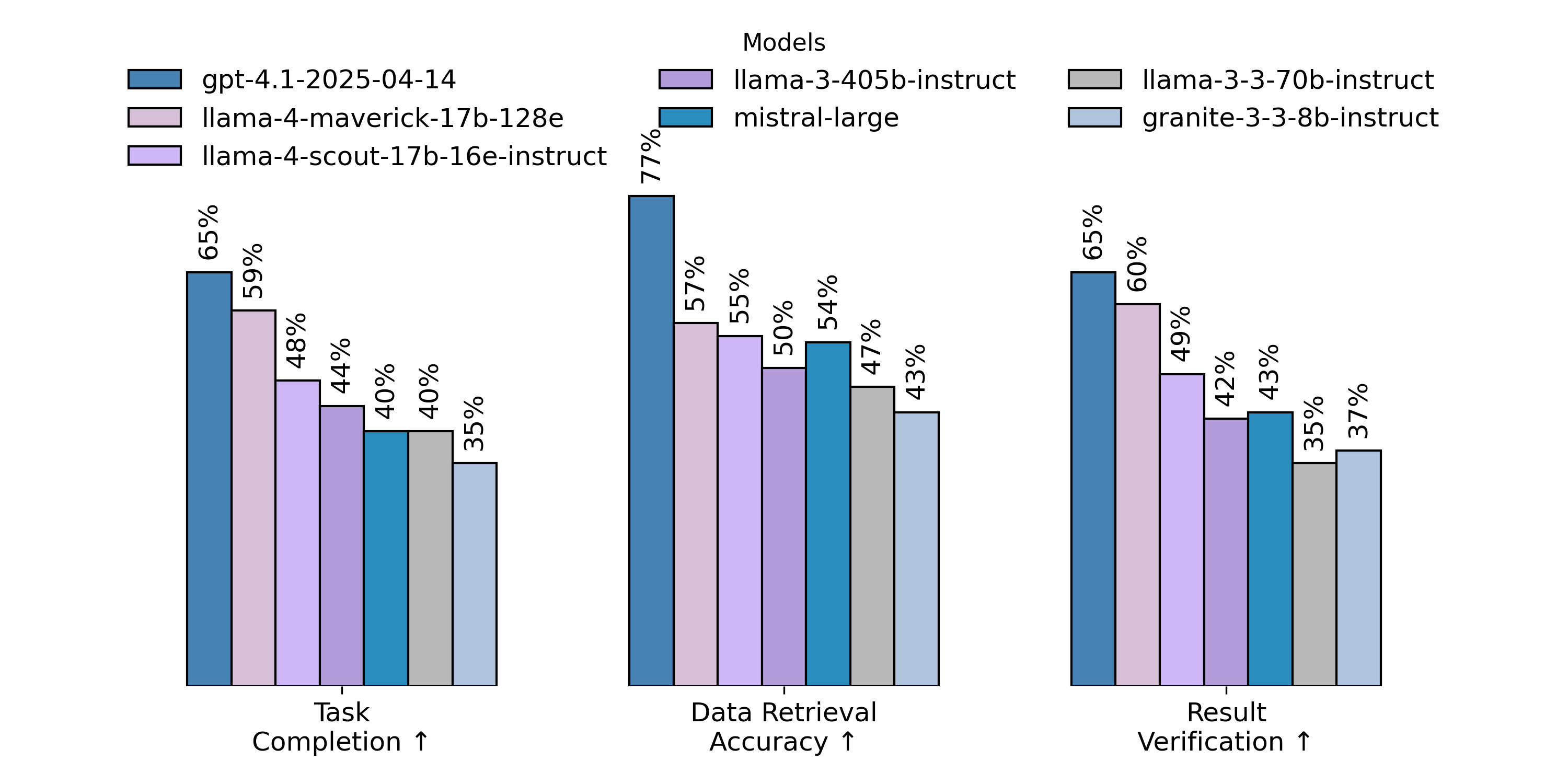}
    \caption{\textbf{Agent-As-Tool} Paradigm}
    \label{fig:agent-as-tool}
\end{subfigure}
\hfill
\begin{subfigure}[t]{0.48\linewidth}
    \centering
    \includegraphics[width=\linewidth]{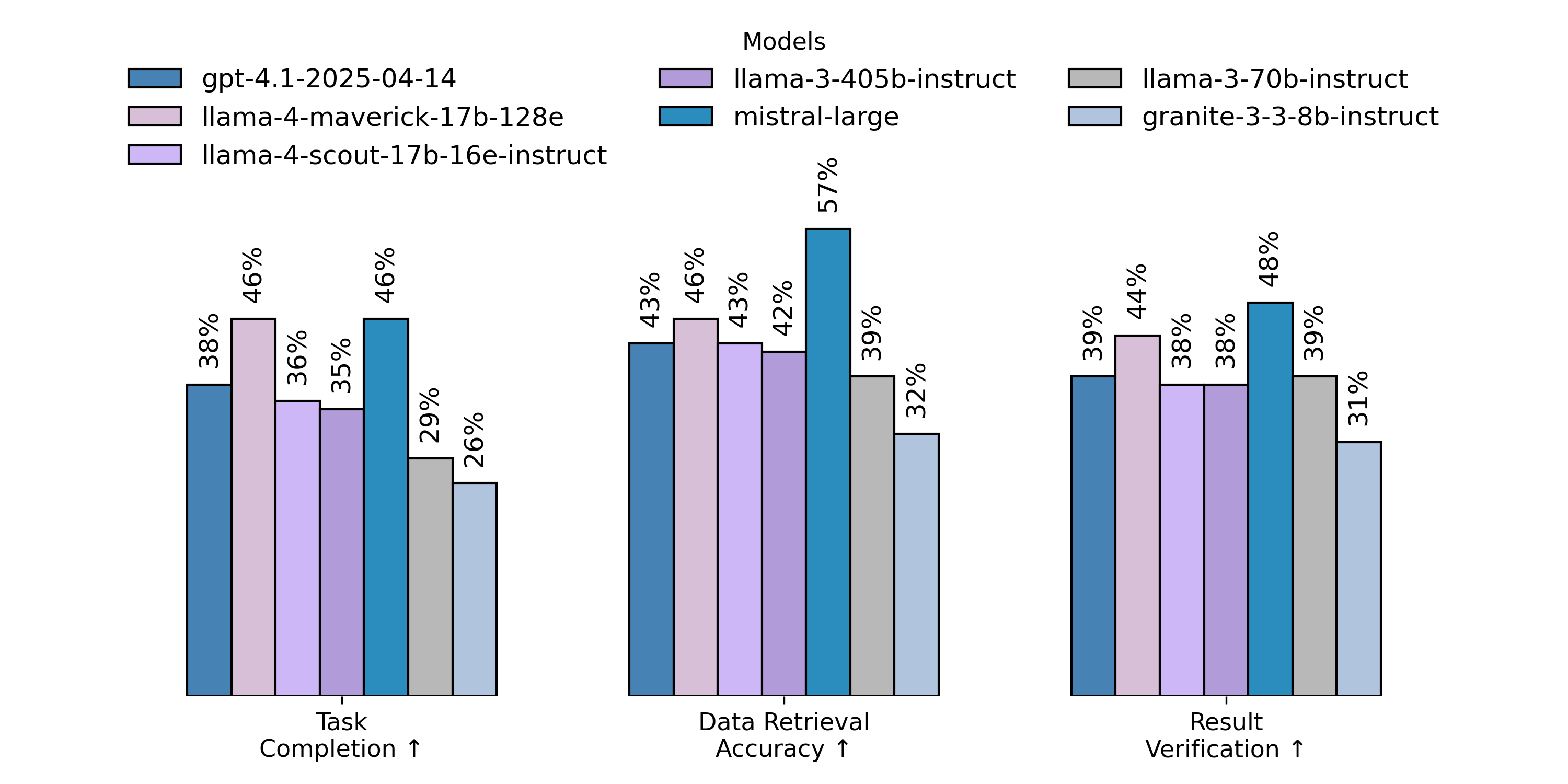}
    \caption{\textbf{Plan-Execute} Paradigm}
    \label{fig:plan-and-execute}
\end{subfigure}
\vspace{-0.1in}
\caption{LLM-As-Judge scoring results across both orchestration systems, ranked by task completion rate ($y_1$).}
\label{fig:combined_performance}
\end{figure*}

\subsection{AssetOpsBench Leaderboard}
\label{leaderboard}

\textbf{Rubric-based Performance Analysis.} 
Figure~\ref{fig:combined_performance} summarizes performance across three dimensions: task completion, data retrieval accuracy, and result verification. \textsc{AssetOpsBench} remains significantly challenging for current LLMs, with no model exceeding a 70\% completion rate. Under the \textbf{Agent-As-Tool} setup (Figure~\ref{fig:agent-as-tool}), \model{gpt\bh{}4.1} leads with the {\color{blue}highest task completion (65\%) and retrieval accuracy (77\%)}. Among open-source candidates, \model{llama\bh{}4\bh{}maverick} is the most competitive, matching the leading models in reasoning stability.

In contrast, \textbf{Plan-Execute} paradigm (Figure~\ref{fig:plan-and-execute}) exhibits a significant performance regression. The peak completion rate drops to 46\%, suggesting that the ``plan-ahead'' pattern introduces cascading failure modes in industrial contexts. Most strikingly, \model{gpt-4.1} shows the sharpest decline ({\color{red} 65\%$\to$38\% completion}), indicating that its proficiency in iterative tool use does not generalize to rigid planning interfaces. Conversely, \model{mistral\bh{}large} and \model{llama\bh{}4\bh{}maverick} emerge as the most robust for Plan-Execute workflows: \model{mistral\bh{}large} achieves the highest retrieval (57\%) and verification (48\%) scores, while \model{llama\bh{}4\bh{}maverick} ties for the top completion rate. Due to its balanced performance across both paradigms, we adopt \model{llama\bh{}4\bh{}maverick} as the default LLM for our ablation studies.

\textbf{Reference-Based Alignment.} 
As shown in Table~\ref{tab:rouge_both_systems}, reference-based scoring reveals a discrepancy between reasoning quality and lexical fidelity. For \textbf{Plan-Execute}, open-weight models like \texttt{mistral-large} and \texttt{llama-3-405b} achieve strong sequence-level alignment ({\color{blue}\texttt{ROUGE-L} $\approx$ 0.34}). However, in the \textbf{Agent-As-Tool} setting, ROUGE scores are heavily influenced by trajectory length. While \model{gpt-4.1} excels in qualitative reasoning, its verbose traces result in lower lexical overlap compared to the concise references. This highlight's a critical limitation: ROUGE captures \textit{textual fidelity} rather than semantic validity. Consequently, reference-based metrics should be interpreted as measures of workflow adherence rather than absolute diagnostic accuracy.

\begin{table*}[t]
\centering
\small
\caption{ROUGE-based reference scoring across both orchestration systems. Plan-Execute scores are computed on planner outputs; Agent-as-Tool scores are computed on extracted thinking traces, where verbosity can reduce ROUGE.}
\label{tab:rouge_both_systems}
\setlength{\tabcolsep}{4.5pt}
\renewcommand{\arraystretch}{1.05}
\begin{tabular}{lcccc c cccc}
\toprule
& \multicolumn{4}{c}{\textbf{Plan-Execute}} && \multicolumn{4}{c}{\textbf{Agent-As-Tool}} \\
\cmidrule(lr){2-5}\cmidrule(lr){7-10}
\textbf{Model} &
\textbf{R-1} & \textbf{R-2} & \textbf{R-L} & \textbf{R-Lsum} &&
\textbf{R-1} & \textbf{R-2} & \textbf{R-L} & \textbf{R-Lsum} \\
\midrule
\rowcolor{highlightgray} 
\texttt{mistral-large} &
\textbf{0.420} & \textbf{0.251} & \textbf{0.343} & \textbf{0.404} &&
\textbf{0.3691} & 0.1933 & \textbf{0.2971} & 0.3124 \\
\texttt{llama-3-405b-instruct} &
0.406 & 0.243 & 0.337 & 0.381 &&
0.3394 & 0.1673 & 0.2740 & 0.2787 \\
\rowcolor{highlightgray}
\texttt{llama-4-maverick} &
0.403 & 0.240 & 0.325 & 0.383 &&
0.2560 & 0.1252 & 0.2067 & 0.2273 \\
\texttt{llama-4-scout-17b-16e-instruct} &
0.395 & 0.227 & 0.265 & 0.368 &&
0.3126 & 0.1522 & 0.2398 & 0.2621 \\
\texttt{gpt-4.1-2025-04-14} &
0.354 & 0.182 & 0.289 & 0.335 &&
0.1628 & 0.0816 & 0.1332 & 0.1389 \\
\texttt{granite-3-3-8b-instruct} &
0.373 & 0.214 & 0.291 & 0.353 &&
0.2473 & 0.1001 & 0.1867 & 0.2079 \\
\texttt{llama-3-3-70b-instruct} &
0.297 & 0.172 & 0.242 & 0.280 &&
0.3661 & \textbf{0.1963} & \textbf{0.2971} & \textbf{0.3177} \\
\bottomrule
\end{tabular}
\end{table*}

\subsection{Operational Efficiency and Evaluation Reliability}
\label{sec:further-analysis}
\textbf{Execution Efficiency.} Table ~\ref{tab:efficiency_chart} reports the mean number of steps (tool invocations) and wall-clock runtime per task for seven LLMs under two agentic implementations. Overall, \textbf{Plan\bh{}Execute} is consistently more step-efficient: most models complete tasks in $\approx$2.6--4.4 steps, whereas \textbf{Agent\bh{}As\bh{}Tool} typically requires $\approx$4--6+ steps due to its iterative \emph{think--act--observe} loop. However, Plan-Execute exhibits limited sensitivity to task complexity: the step counts for single-agent and multi-agent tasks are often nearly identical (e.g., \texttt{gpt-4.1}: 2.6 vs.\ 2.9; \texttt{granite-3-3-8b}: 5.2 vs.\ 5.1), suggesting a tendency to {\color{red}\emph{over-plan}} even when the objective is relatively simple.

Runtime trends are more nuanced than step counts (See Runtime columns in Tables~\ref{tab:efficiency_chart}). While fewer steps can reduce execution time for some models (e.g., \model{gpt\bh{}4.1} remains the fastest and most step\bh{}efficient Plan\bh{}Execute backbone, especially on multi\bh{}agent tasks), several models incur substantial overhead in Plan\bh{}Execute (e.g., \model{llama\bh{}3\bh{}70b\bh{}instruct} and \model{granite\bh{}3\bh{}3\bh{}8b} exceed $\sim$370--430 seconds on multi-agent tasks). In contrast, Agent-As-Tool can be competitive---and for some backbones, dramatically faster---despite taking more steps (e.g., \texttt{llama-4-maverick} drops from $\sim$377s in Plan-Execute to $\sim$137s in Agent-as-Tool on multi-agent tasks), indicating lower per-step overhead and more direct tool control. At the same time, tool/data effects can dominate end-to-end latency: \texttt{mistral-large} shows high variance in Agent-As-Tool due to a pathological outlier with prolonged \texttt{JSONReader} calls over large histories.

Overall, Plan-Execute offers structural and step efficiency, but its wall-clock cost depends strongly on model--planner--executor alignment, whereas Agent-As-Tool often spends more steps yet tends to deliver stronger end-task quality when correctness is prioritized.

\begin{table*}[t]
\centering
\small
\caption{Execution efficiency summary: mean steps per tasks and mean runtime per tasks. Single-agent tasks contain 99 utterances, while multi-agent tasks contain 42 utterances.}
\label{tab:efficiency_chart}

\setlength{\tabcolsep}{4pt}
\renewcommand{\arraystretch}{1.05}

\begin{tabular}{lcccc c cccc}
\toprule
& \multicolumn{4}{c}{\textbf{Agent-As-Tool}} && \multicolumn{4}{c}{\textbf{Plan-Execute}} \\
\cmidrule(lr){2-5}\cmidrule(lr){7-10}

& \multicolumn{2}{c}{\textbf{Single-agent tasks}} & \multicolumn{2}{c}{\textbf{Multi-agent tasks}} &&
  \multicolumn{2}{c}{\textbf{Single-agent tasks}} & \multicolumn{2}{c}{\textbf{Multi-agent tasks}} \\
\cmidrule(lr){2-3}\cmidrule(lr){4-5}
\cmidrule(lr){7-8}\cmidrule(lr){9-10}

\textbf{Model} &
\textbf{Steps} & \textbf{Runtime (Sec)} &
\textbf{Steps} & \textbf{Runtime (Sec)} &&
\textbf{Steps} & \textbf{Runtime (Sec)} &
\textbf{Steps} & \textbf{Runtime (Sec)} \\
\midrule

\texttt{gpt-4.1} & 6.0 & 104 & 6.4 & 218 && 2.6 & 93 & 2.9 & 180 \\
\texttt{mistral-large} & 4.9 & 347 & 5.2 & 289 && 2.7 & 187 & 3.0 & 210 \\
\texttt{llama-3-405b-instruct} & 4.8 & 250 & 5.6 & 255 && 3.1 & 208 & 4.0 & 224 \\
\texttt{llama-3-70b-instruct} & 3.9 & 101 & 4.3 & 151 && 6.7 & 382 & 6.5 & 370 \\
\texttt{llama-4-maverick-17b-128e} & 4.3 & 120 & 4.5 & 137 && 4.0 & 385 & 3.9 & 377 \\
\texttt{llama-4-scout-17b-16e} & 4.4 & 101 & 5.8 & 178 && 3.9 & 172 & 4.4 & 218 \\
\texttt{granite-3-3-8b} & 5.3 & 197 & 6.6 & 228 && 5.2 & 413 & 5.1 & 433 \\
\bottomrule
\end{tabular}
\end{table*}

\begin{figure*}[t!]
  \centering
  \begin{subfigure}[t]{0.35\textwidth}
    \centering
    \includegraphics[width=\linewidth]{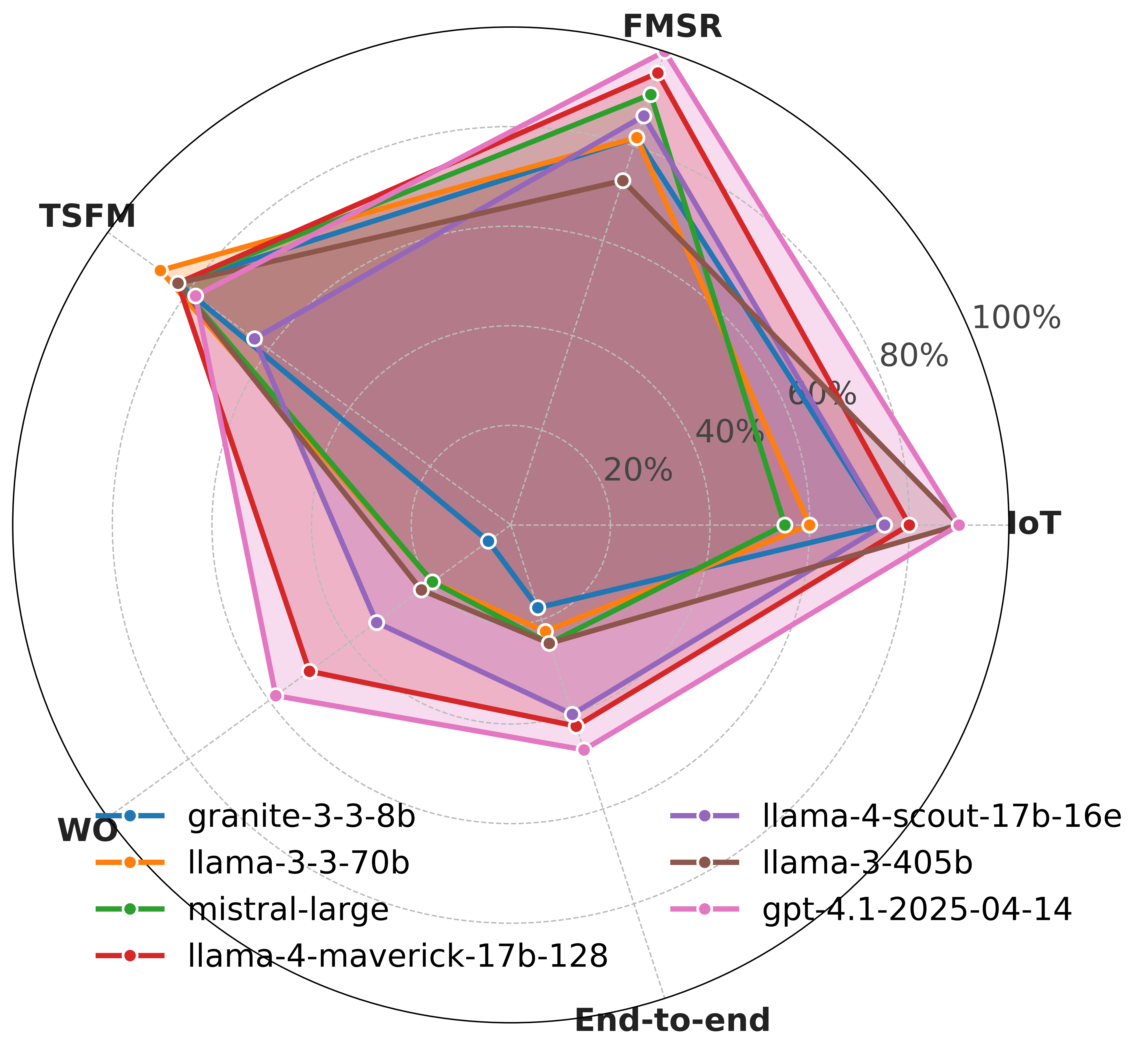}
    \caption{Agent-level task accomplishment w.r.t. Agent-As-Tool approach.}
    \label{fig:model_radar_chart}
  \end{subfigure}
  \begin{subfigure}[t]{0.64\textwidth}
    \centering
    \includegraphics[width=\linewidth]{images/fm_distribution_layout.png}
    \caption{Distribution of failure modes over 881 selected agent trajectories.}
    \label{fig:fm_dis}
  \end{subfigure}
  \caption{Agent performance and failure-mode coverage in AssetOpsBench.}
  \label{fig:combined_eval}
\end{figure*}

\textbf{Small Language Models (SLM) Analysis.} Overall, SLMs tend to underperform. However, we conduct a more detailed analysis to identify where SLMs are effective and where they break down. In the Agent-As-Tool evaluation, models such as \texttt{granite-3-8b} and \texttt{llama-3-3-70b} exhibit weaker aggregate performance, yet they demonstrate clear areas of specialization, as shown in Figure \ref{fig:model_radar_chart}. Both models perform strongly on structured sensing and diagnostic tasks: for example, \texttt{granite-3-3-8b} achieves 15/20 on IoT, 18/22 on FMSR, and 19/23 on TSFM, while \texttt{llama-3-3-70b} reaches 12/20, 18/22, and 20/23 on the same categories. However, they struggle substantially on Work Order tasks, with scores of only 2/36 and 7/36, indicating that procedural, multi-step coordination remains difficult even under the Agent-As-Tool mechanism. This underscores a key insight: industrial deployments may benefit most from \textbf{hybrid LLM–SLM agent architectures}, where strong specialists handle sensing and diagnostics while more capable generalist models manage planning and end-to-end coordination \citep{belcak2025smalllanguagemodelsfuture}.

\textbf{Human Validation of LLM Judge.} To assess the reliability of using LLMs as automatic evaluators for benchmarking tasks, we compare model-generated judgments against human annotations on a sample of 40 tasks. Each task is evaluated along three dimensions by four domain experts, all operating under the same information constraints as the LLMs. {\color{black}Before selecting a default evaluator, we compared several candidate judge models, including \texttt{gpt-4.1}. In this comparison, \texttt{gpt-4.1} showed only moderate alignment with expert assessments, achieving 69\% accuracy and Cohen’s $\kappa$ of 0.44}. In contrast, \texttt{llama-4-maverick} provided substantially stronger agreement with human judgments and was therefore selected as the default judge model for the main analysis. Across experts, inter-rater reliability scores indicate substantial agreement on key evaluation dimensions, with \textit{Data Retrieval Accuracy} exhibiting the strongest consistency (Cohen's $\kappa$ = 0.79, 90.48\% accuracy). \textit{Task Completion} ($\kappa$ = 0.62) and \textit{Generalized Result Verification} ($\kappa$ = 0.71) also show high alignment among evaluators.

See Appendix~\ref{astudy} for ablation studies on distractor agent impact and single- versus multi-agent performance comparisons.

\subsection{Scenarios Realism Analysis}
To validate the operational authenticity of \textsc{AssetOpsBench}, we conducted a formal expert validation study. A panel of domain experts—including reliability engineers, maintenance specialists, and data scientists—evaluated a random sample of 25 representative scenarios. As detailed in Appendix Section \ref{sec:reliability}, the results demonstrate high statistical robustness. The study yielded a \textbf{Cronbach's Alpha of 0.89}, confirming strong internal consistency across the scenario ratings. Although we observed natural variability of individual raters ($\text{ICC(1)} = 0.23$), aggregate reliability was high (\textbf{$\text{ICC(2)} = 0.88$}), indicating that averaged expert judgments provide a highly stable estimate of perceived realism. Furthermore, a Fleiss' Kappa of 0.21 was obtained; while seemingly modest, this level of agreement is consistent with subjective realism assessments in complex, high-stakes industrial domains. These findings verify that our 141 scenarios faithfully encapsulate the diagnostic reasoning and technical rigor required in professional industrial practice.

\subsection{Failure Analysis via Agent Trajectories}
We analyze agent reasoning traces (trajectories) to identify systemic failure modes. Across approximately 881 collected trajectories for the Agent-As-Tool strategy, we conduct a two-pronged error analysis focusing on \textbf{tool-related errors} and \textbf{agent failure modes}.

\textbf{Failure Analysis on Tool Use.} Each agent step in a trajectory is logged as a structured JSON record capturing the \emph{action type} and \emph{execution state}. At the sub-agent level, we distinguish between \textbf{Tool-oriented actions}, which invoke predefined functions with well-defined inputs and outputs, and \textbf{CodeReAct-oriented actions}, where agents dynamically generate and execute Python code. Our analysis shows that Tool-oriented actions achieve higher valid-execution rates, whereas CodeReAct-oriented actions incur more runtime failures due to the variability of the generated code. Tool-oriented failures are concentrated in a small number of tools, including \texttt{jsonreader}, \texttt{tsfm\_integrated\_tsad}, and \texttt{Read Sensors From File}, highlighting challenges related to input validation and hallucinated parameter passing.

\textbf{Emerging Failure Modes Discovery.} Now we investigate trajectories from a semantic perspective. Recent work~\citep{cemri2025multi} defines 14 failure modes for agent trajectories. Figure~\ref{fig:fm_dis} shows the distribution of the failure mode in our 881 trajectories in this taxonomy. We found that \textbf{system design} is the most common source of failures. This taxonomy provides guidance to improve agent development. For instance, since the ``Fail to Ask for Clarification'' mode occurs around 10\% of the time, we introduced a feature in the Agent-As-Tool strategy that allows sub-agents to ask the parent agent a clarification questions at any point during execution. We reran the entire benchmark on the default LLM, and this change led to significant performance improvements for \texttt{llama-4-maverick}, increasing task completion from 59\% to 66\%, surpassing \texttt{gpt-4.1}. To capture failure mode behaviors beyond this taxonomy, we allowed self-discovery of up to two \textbf{novel failure modes} per trace, revealing \textit{emergent and compound failures} not covered by existing classifications. This automated diagnostic find several emergent failure modes: \textbf{Overstatement of Task Completion} (122 cases, 23.8\%), \textbf{Extraneous or Ambiguous Output Formatting} (110 cases, 21.4\%) and \textbf{Ineffective Error Recovery} (160 cases). 


\section{Community Adoption and Generalization}
\label{impact}
\textbf{Community Engagement.} To move beyond static evaluation \cite{huggingface_community_evals_2026}, we operationalized \textsc{AssetOpsBench} as a live, code-executed competition on \textbf{Codabench} \cite{codabench2025}. This platform enables end users to integrate custom agent architectures and evaluate them against hidden scenarios via containerized submissions. Over a three-month period, the competition engaged 250+ participants (44.2\% Undergraduate, 28.4\% Industry, 23.6\% Postgraduate, and 3.8\% Other), resulting in 500+ agent submissions across specialized planning and execution tracks. This deployment demonstrates the system's capability to host custom agents and maintain a dynamic leaderboard; furthermore, to maximize academic impact and inclusivity, we provided full LLM API support funded through our internal budget to ensure participants could iterate without financial barriers.

\begin{figure}
    \centering
\includegraphics[width=0.9\linewidth]{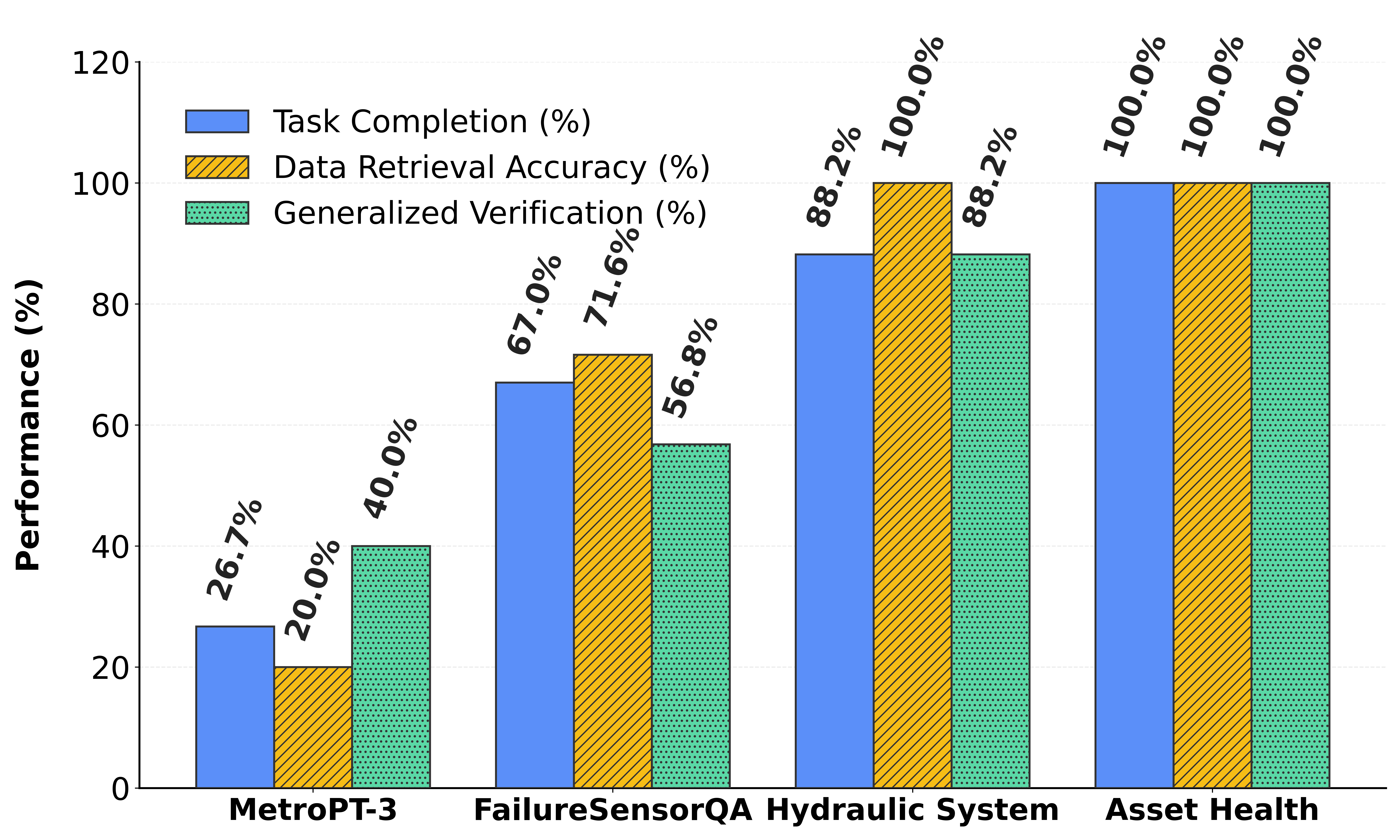}
    \caption{Extended Validation for Different Asset Class}
    \vspace{-0.15in}
    \label{fig:extvalidation}
    \vspace{-0.15in}
\end{figure}

\textbf{Generalization Across Industrial Assets.} To evaluate generalization capabilities, we developed 162 additional scenarios across four distinct datasets and an extended environment (Figure \ref{fig:env}) with new datasets and scenarios. These datasets include MetroPT-3 \cite{metropt-3_dataset_791} (15 scenarios) focused on compressor faults, the UCI Hydraulic System \cite{condition_monitoring_of_hydraulic_systems_447}(17 scenarios) for hydraulic component failures, an internal Asset Health dataset (42 scenarios) derived from industrial work orders, and FailureSensorQA (88 scenarios) based on ISO-standardized sensor-to-failure mapping. Figure \ref{fig:extvalidation} shows summary of results for Agent-As-Tool approach. Among all the datasets, scenarios of MetroPT-3 are difficult as we observed poor performance (task completion rate = 26.7\%), whereas Asset Health (a production usecase) had a near perfect solution.

\section{Conclusion and Future Outlook}
This paper introduces \textsc{AssetOpsBench}, a formalized framework for industrial AI agents using a standardized evaluation methodology and the Agent-As-Tool paradigm to orchestrate complex multi-agent interactions. Through a competition that engaged 250+ participants across academia and industry, we demonstrated the system's capability to host custom agents and foster iterative refinement via reproducible, code-executed evaluations. Future extensions will incorporate realistic constraints, such as compute limits and API costs, to drive the development of cost-aware autonomous industrial algorithms.

\bibliographystyle{ACM-Reference-Format}
\bibliography{sample-kdd}


\appendix
\section{Details on Multi-Agent Systems}
\label{sec:appendix_agents}

AssetOpsBench uses an extended ReAct framework \cite{rayfield-etal-2025-react} as the base agent to both multi-agent systems. Subsequent sections present the architecture in details, highlighting the distinction between two architectural paradigms: Agent-As-Tool (See \ref{sec:aat}) and Plan-Execute (See \ref{sec:pe}). Note that, we can replace ReAct by
any other agent development methdology such as Reflect, RAFA, etc.

\begin{figure*}[t]
    \centering
    \includegraphics[width=.8\linewidth]{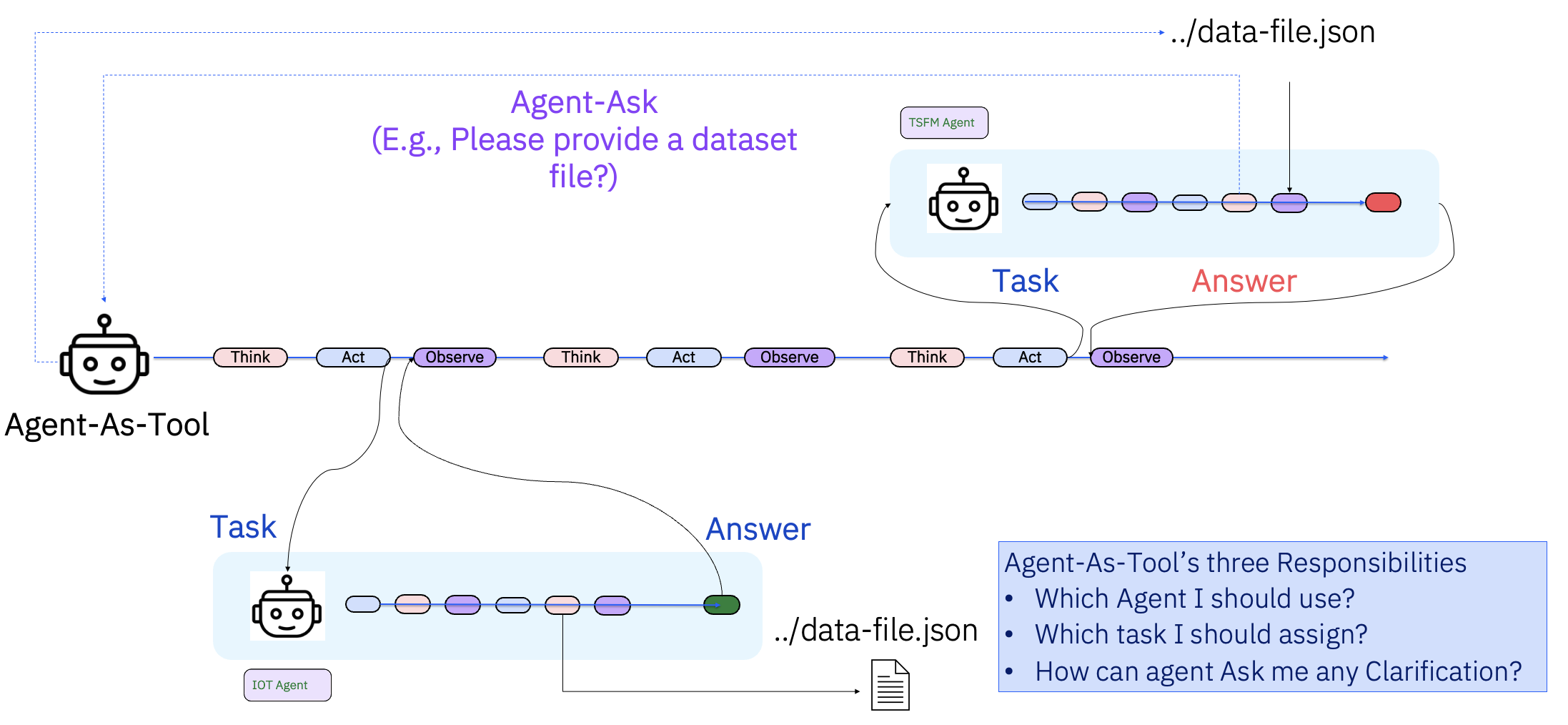}
    \caption{Typical Workflow: Agent-As-Tool using Agent-Ask.}
    \label{fig:agentastoolworkflow}
\end{figure*}

\subsection{Agent As Tools}
\label{sec:aat}
An Agent-as-Tool multi-agent system is a multi-agent setup where each agent is treated like a callable tool, meaning agents are not fully autonomous “actors,” but modular capabilities that can be invoked on demand by an orchestrator agent or a meta agent. 

The Agent-As-Tool paradigm, coupled with the Agent-Ask mechanism, offers a robust framework for complex task execution by decomposing a problem and routing sub-tasks to specialized AI agents. As illustrated in the workflow Figure \ref{fig:agentastoolworkflow}, the primary Agent-As-Tool receive a user query. The Agent-As-Tool has to fulfill three core responsibilities: (1) determining the appropriate specialized agent (e.g., the TSFM Agent or IoT Agent) for a given sub-task, (2) correctly assigning the task, and (3) facilitating clarification from the specialized agent if needed. Once the Agent-As-Tool receives an input query, it enters a standard Think-Act-Observe loop, and it decides on the appropriate specialized agent in think step. The specialized agent executes its designated task and provides an Answer. Answer is embedded inside to the orchestration agent's observation for next set of action. In this case, an artifact like a data-file.json, back to the main orchestrator, allowing the overall system to complete complex, multi-faceted operations that exceed the capability of any single agent.

Algorithm \ref{alg:agent-as-tool} details an iterative, multi-turn execution framework akin to the ReAct (Reasoning and Acting) paradigm. This framework enables a central agent to solve a complex User Task ($\mathcal{Q}$) by strategically engaging specialized agents ($\mathcal{A}_{spec}$) within a bounded number of steps $T_{max}$. The process begins with the initialization of the agent's internal state $M$ and an empty execution history $\mathcal{H}$. The agent then enters an iterative loop where a thinking policy $\Pi_{think}(M)$ determines the next $\text{action}$ (either $\text{THINK-ACT}$ or $\text{FINISH}$) and the continuation signal. When the action is $\text{THINK-ACT}$, the agent selects the best $\text{agent\_id}$ via $\Pi_{select}$ and formulates a precise $\text{sub\_task}$ via $\Pi_{formulate}$ for execution. Conversely, if the action is $\text{FINISH}$, the agent summarizes the full $\Pi$, updates the memory, and terminates the loop. Following any action, the resulting $\text{output}$ is compressed into a concise $\text{observation}$ ($\Pi_{compress}$) to manage context length. This observation is logged to $\mathcal{H}$, and critically, the agent's internal memory $M$ is updated with this new context, driving the decision-making in the subsequent round. Finally, once the loop terminates, a policy $\Pi_{final}$ generates the complete $\text{Final Output}$ ($O$) by synthesizing the entire execution history $\mathcal{H}$.

To implement Agent-As-Tools, we implemented the following components:

\begin{itemize}
\item A standard ReAct (Think–Act–Observe) agent loop using open source framework. In the initial setup, the \textit{number of reflections} was set to one—effectively disabling reflection. 
\item A curated list of tools, the majority of which are stub interfaces that delegate functionality to specialized sub-agents. The only standalone utility tool in this set was the \texttt{JSONReader}, which reads a JSON object from a file and returns its contents as the tool's direct response.
\end{itemize}

The sub-agent stubs were intentionally designed to be minimal. Each stub accepted a single input parameter: a string called \texttt{request} and returned a structured JSON output. The output JSON object included the following fields:

\begin{itemize}
\item \texttt{answer} – the primary answer returned by the sub-agent, represented as a plain string.
\item \texttt{review} – a nested JSON object capturing a review of the response, typically including fields such as \texttt{status}, \texttt{reasoning}, and \texttt{suggestions}.
\item \texttt{summary} – a brief description of the JSON object's structure and semantics, useful for interpretability or chaining with downstream tools.
\end{itemize}

\begin{algorithm}[t]
\small
\caption{Agent-As-Tool (Simplified as ReAct)}
\label{alg:agent-as-tool}
\SetKwInOut{Input}{Input}
\SetKwInOut{Output}{Output}
\SetKwComment{Comment}{//}{}

\Input{User Query $\mathcal{Q} \in \mathcal{Q}$, Maximum Steps $T_{max}$, Set of Agents $\mathcal{A}_{spec} \subseteq \mathcal{A}$}
\Output{Final Output $O \in \mathcal{O}$, Execution Plan $\Pi$}

\BlankLine
$M \leftarrow \text{InitializeMemory}(\mathcal{Q})$ \Comment{Initialize the global Memory System $M$}
$\Pi \leftarrow \emptyset$\;
$\mathcal{H} \leftarrow \emptyset$\;
\For{$t = 1$ \KwTo $T_{max}$}{
    $(continue, \text{action}) \leftarrow \Pi_{think}(M)$ \Comment{Agent decides next step using current Memory $M$}
    
    \If{$\text{action} = \text{THINK-ACT}$}{
        $\text{agent\_id} \leftarrow \Pi_{select}(\text{action}, M, \mathcal{A}_{spec})$ \Comment{Select best Agent $A_i \in \mathcal{A}_{spec}$}
        $\tau \leftarrow \Pi_{formulate}(\text{action}, M)$ \Comment{Formulate sub-task $\tau \in \mathcal{T}$ based on current Memory $M$}
        $\text{output} \leftarrow \text{ExecuteAgent}(\text{agent\_id}, \tau)$ \Comment{Agent executes task and returns structured output $o \in \mathcal{O}$}
        $\Pi \leftarrow \Pi \cup \{\langle \tau, \text{agent\_id} \rangle\}$ \Comment{Update the execution plan $\Pi$ with task-agent assignment}
    }
    \ElseIf{$\text{action} = \text{FINISH}$}{
        $\text{final\_output} \leftarrow \text{Summarize}(\Pi)$\;
        $M \leftarrow \text{UpdateMemory}(M, \text{final\_output})$\;
        \textbf{break}\;
    }
    
    $\text{observation} \leftarrow \Pi_{compress}(\text{output})$ \Comment{Generate concise observation summary (optional}
    $\mathcal{H} \leftarrow \mathcal{H} \cup \{(t, \text{action}, \text{observation})\}$ \Comment{Log step to history}
    $M \leftarrow \text{UpdateMemory}(M, \text{observation})$ \Comment{Update Memory $M$ with new context from observation}
    
    \If{$continue = \text{False}$}{
        \textbf{break} \Comment{Stop if agent decides termination}
    }
}
$O \leftarrow \Pi_{final}(\mathcal{H})$ \Comment{Produce final answer from the complete history}
\Return $(O, \Pi, \mathcal{H})$
\end{algorithm}

\begin{figure}[ht]
\centering
\begin{minipage}{\linewidth}
\lstset{
  basicstyle=\ttfamily\footnotesize, 
  breaklines=true,                   
  postbreak=\mbox{\textcolor{red}{$\hookrightarrow$}\space}, 
  frame=single,                      
  captionpos=b                       
}
\begin{lstlisting}[language=,caption={Example of Trajectory calling IoTAgent in Agent-As-Tool workflow},label={lst:react_iot_example}]
Question: download asset history for CU02004 at SiteX 
from 2016-07-14T20:30:00-04:00 to 2016-07-14T23:30:00-04:00 
for "CHILLED WATER LEAVING TEMP" and 
"CHILLED WATER RETURN TEMP"

Action 1: IoTAgent
Action Input 1: request=download asset history for CU02004 
at SiteX from 2016-07-14T20:30:00-04:00 to 
2016-07-14T23:30:00-04:00 for "CHILLED WATER LEAVING TEMP" 
and "CHILLED WATER RETURN TEMP"

Observation 1: {
  "site_name": "SiteX",
  "assetnum": "CU02004",
  "total_observations": 25,
  "start": "2025-03-26T00:00:00.000000+00:00",
  "final": "2025-04-02T00:00:00.000000+00:00",
  "file_path": "/var/folders/fz/.../cbmdir/c328516a-643f-40e6-8701-e875b1985c38.json",
  "message": "found 25 observations. file_path contains a JSON array of Observation data"
}
\end{lstlisting}
\end{minipage}
\end{figure}

We have provided a sample (partial) trajectory trace in Listing \ref{lst:react_iot_example}, which show how patent agent call one of the tool (in this case IoTAgent) and receive a response. The recorded metrics include:

\begin{itemize}
    \item \textbf{Question:} the input query being processed
    \item \textbf{Total execution time:} duration of the entire ReAct loop
    \item \textbf{Number of ReAct steps:} count of action-observation cycles
    \item \textbf{Review status:} success or failure determined by the LLM-based reviewer
\end{itemize}

\subsection{Plan Execute}
\label{sec:pe}
\begin{figure}[h!]
    \centering
    \includegraphics[width=0.8\linewidth]{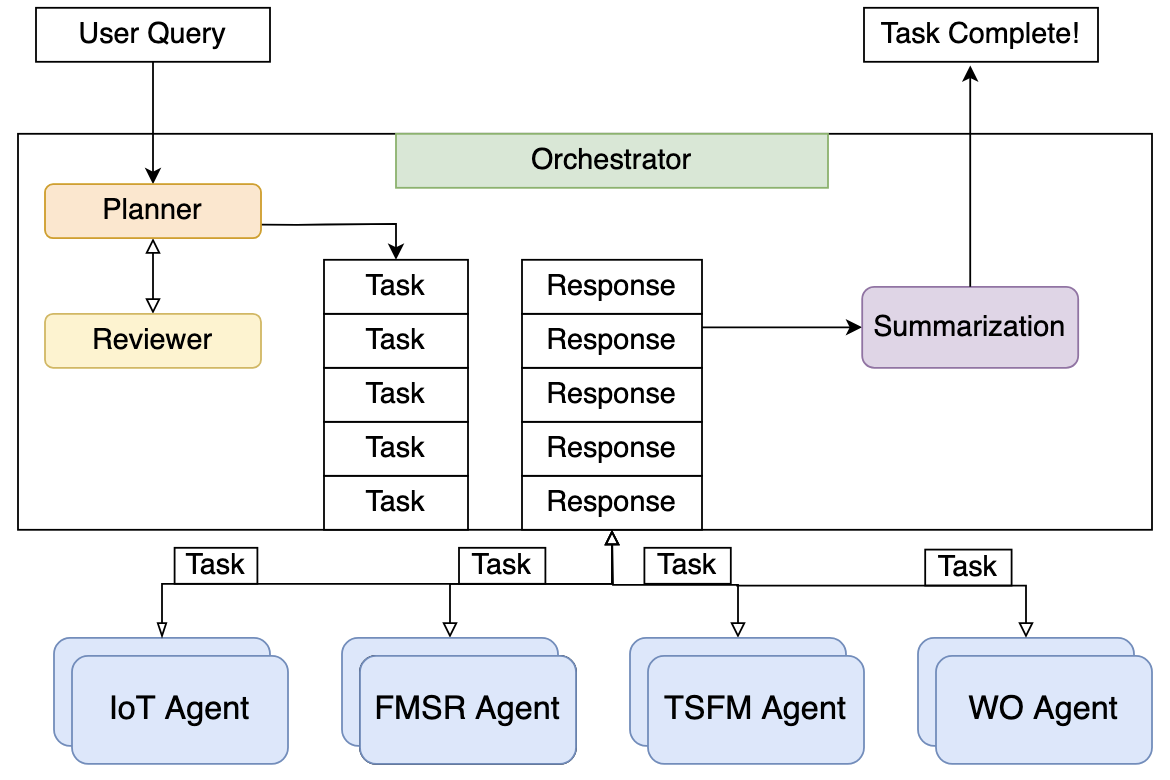}
    \caption{Plan-Execute Multi-Agent System}
    \label{fig:Plan-Execute-diagram}
\end{figure}

Plan-Execute is a widely used architectural paradigm for multi-agent systems. Figure~\ref{fig:Plan-Execute-diagram} depicts the implementation adopted in our work. It is derived from specialized multi-agent system \cite{marreed2025enterprisereadycomputerusinggeneralist}. The process initiates when a user submits a query, which is first processed by the \textbf{Planner}. The Planner decomposes the query into discrete, executable tasks. These tasks are then vetted by a \textbf{Reviewer} component to ensure quality, completeness, and relevance. Upon approval, the \textbf{Orchestrator} assigns the tasks to the most appropriate agents. Each agent independently executes its assigned task and returns a structured response. These responses are then aggregated by the \textbf{Summarization} module, which synthesizes them into a coherent final output that is returned to the user. 

This architecture supports modularity, robustness, and interpretability across the task lifecycle. We have provided two system prompts where first prompt guides an AI to generate a structured step-by-step plan using external agents, while the second prompt instructs a reviewer agent to evaluate the plan's correctness and completeness in JSON format.

\begin{tcolorbox}[title=System Prompt (Planning Agent),
                  label={lst:planning_prompt},
                  colback=blue!3,
                  colframe=blue!50!black,
                  breakable]

\begin{lstlisting}[language=Python, breaklines=true]
You are an AI assistant who makes step-by-step plan to solve a complicated problem under the help of external agents. 
For each step, make one task followed by one agent-call.
Each step denoted by #S1, #S2, #S3 ... can be referred to in later steps as a dependency.

Each step must contain Task, Agent, Dependency and ExpectedOutput. 
1. **Task**: A detailed description of what needs to be done in this step. It should include all necessary details and requirements.
2. **Agent**: The external agent to be used for solving this task. Agent needs to be selected from the available agents.
3. **Dependency**: A list of previous steps (denoted as #S1, #S2, etc.) that this step depends on. If no previous steps are required, use None.
4. **ExpectedOutput**: The anticipated result from the agent's execution.

## Output Format (Replace '<...>') ##

## Step 1
#Task1: <describe your task here>
#Agent1: <agent_name>
#Dependency1: None
#ExpectedOutput1: <describe the expected output of the call>

## Step 2
#Task2: <describe next task>
#Agent2: <agent_name>
#Dependency2: [#S1]
#ExpectedOutput2: <describe the expected output of the call>

And so on...

Here are the available agents:
{agent_descriptions}
 
You are going to solve the following complicated problem:
{task.description}

Guidelines:
- Task should be something that can be solved by the agent. 
- A plan usually contains less than 5 steps.
- Only output the generated plan.

Output (your generated plan):
\end{lstlisting}
\end{tcolorbox}

\begin{tcolorbox}[title=System Prompt (Review Agent),
                  colback=blue!3,
                  colframe=blue!50!black,
                  breakable]

\begin{lstlisting}[basicstyle=\footnotesize\ttfamily, breaklines=true]
review_plan_system_prompt_template = """You are a critical reviewer tasked with evaluating the effectiveness and accuracy of a plan. Your goal is to determine whether the plan is valid or not given the context of the input question and agent expertise. A valid plan should:

1. **Ensure all necessary actions are addressed:** The plan must cover all required steps to successfully complete the task as specified in the question. Ensure that each action directly contributes to the task goal.
2. **Include appropriate dependencies between steps:** Actions should be logically ordered with clear dependencies. Each step must rely on the completion of the previous step to ensure a coherent and efficient workflow.
3. **Ensure no crucial steps are missed:** The plan must not overlook any essential actions required to solve the task. If any crucial steps are absent, the plan must be flagged as incomplete.
4. **Confirm all actions align with agent capabilities:** Each step in the plan must fall within the designated expertise of the agents involved. No action should require expertise or knowledge outside of the agent's specified capabilities. Any plan that violate this condition is an invalid plan.
5. **Strictly follow the task's question:** Carefully compare the provided question with the task. The plan should only include actions that directly relate to the question's explicit requirements, without introducing any unnecessary tasks or assumptions.
6. **Avoid Abstract task/step:**
   Ensure steps/tasks are grounded with respect to the data generated by previous steps or the question.

### Evaluation Criteria:
1. **Completeness:**
   - Verify that the system prompt leads to a plan that includes all necessary steps to accomplish the task.
   - Ensure the description of each step contains all the relevant information needed to execute the step, including any required parameters or inputs that are mentioned in the task's question.

2. **Relevance:**
   - Confirm that each step in the plan directly contributes to solving the task.
   - Eliminate any steps that do not serve a clear purpose in achieving the goal.

3. **Correctness:**
   - Ensure that all steps are logically consistent and ordered correctly.
   - Ensure that the dependencies between the steps are valid and follow a correct sequence.

4. **Expertise Alignment:**
   - Confirm that the steps in the plan are within the capabilities of the agent.
   - Validate that the agents used in each steps are among the available agents mentioned in the agents' expertise.
   
5. **Efficiency:**
   - Make sure the plan doesn't introduce redundant actions.
   - Avoid unnecessary complexity in the plan.

6. **Clarity:**
   - Ensure that the plan is easy to understand and logically structured.

---

**Question:** {question}

**Agents' Expertise:**
{agent_expertise}

**Plan:** {plan}

---

### Output Format:
Your review must always be in JSON format. Do not include any additional formatting or Markdown in your response.

```json
{{
    "status": "Valid | Invalid | Other",
    "reasoning": "A concise explanation for your evaluation. If a specific step is wrong, point it out directly.",
    "suggestions": "Actions or improvements for rectifying the plan if applicable."
}}
```

Output:
"""
\end{lstlisting}
\end{tcolorbox}

\begin{figure*}[t!]
    \centering
    \includegraphics[width=0.9\linewidth]{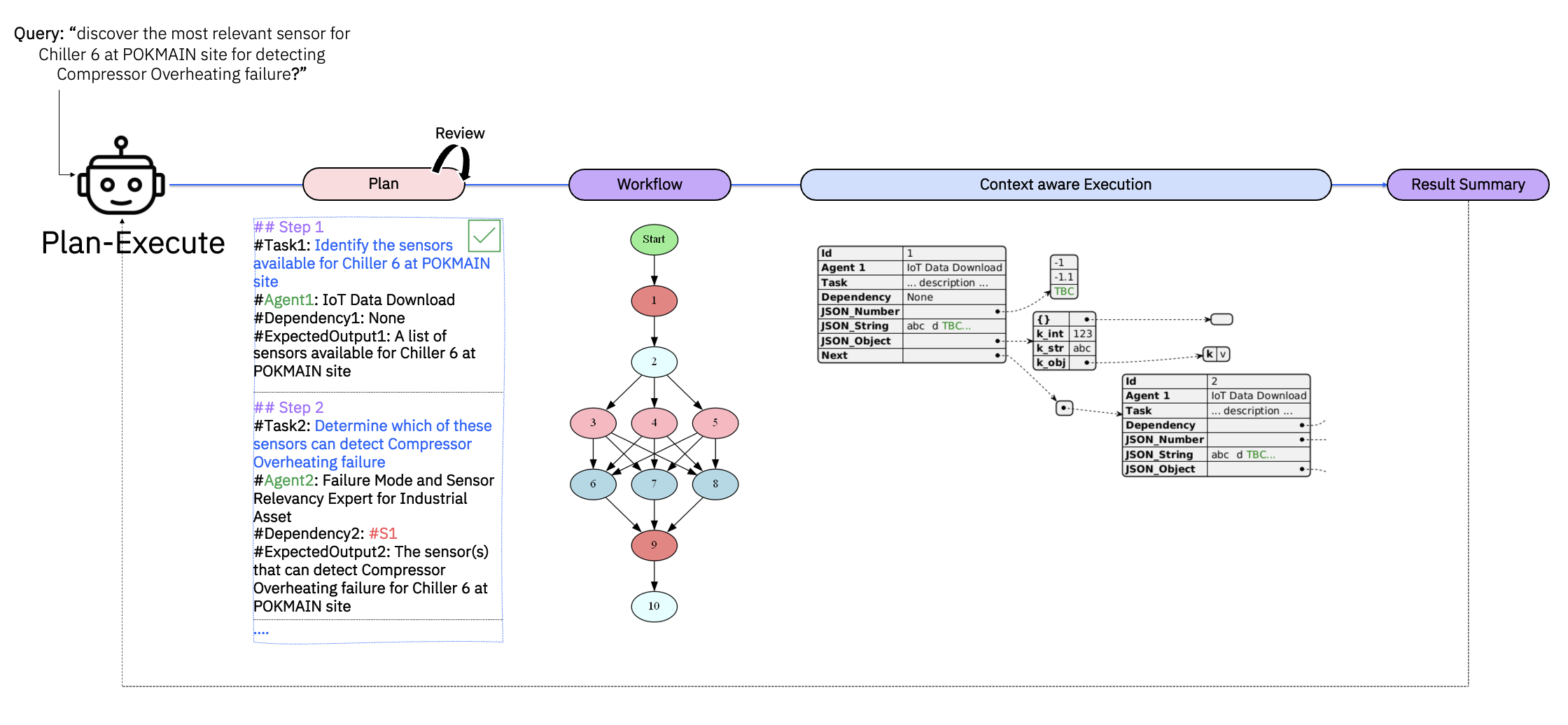}
    \caption{Plan-Execute Approach using Review}
    \label{fig:planworkflow}
\end{figure*}

\begin{figure*}[h!]
    \centering
    \includegraphics[width=0.9\linewidth]{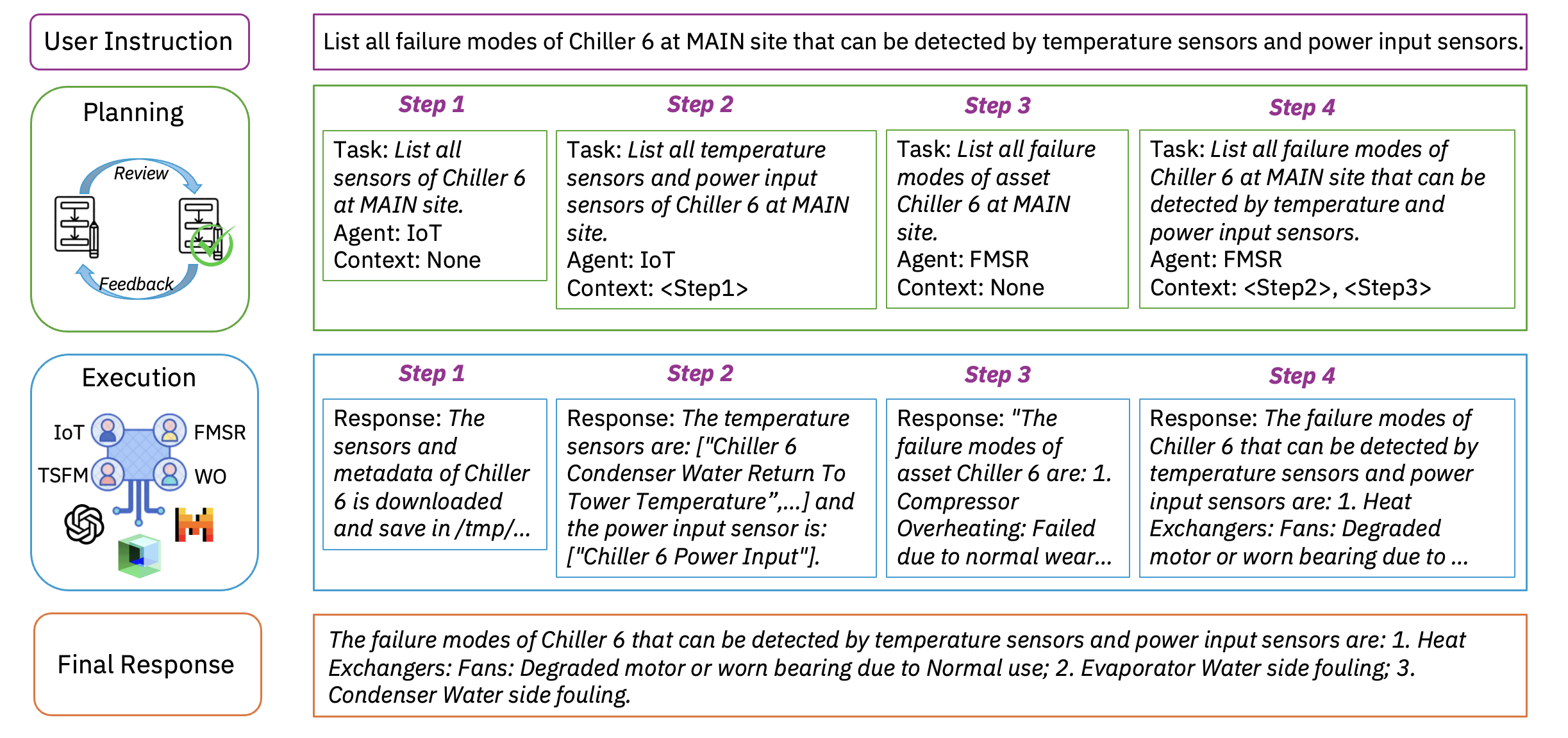}
    \caption{Plan-Execution Workflow Stepwise Concrete Example}
    \label{fig:pe-illustration}
\end{figure*}

Figure \ref{fig:planworkflow} and Figure \ref{fig:pe-illustration} illustrate the Plan-Execute approach to tackle a complex industrial query, such as ``discover the most relevant sensor for Chiller 6 at POKMAIN site for detecting Compressor Overheating failure?''. The process begins with the main agent receiving the Query and formulating a detailed Plan. This plan is meticulously broken down into sequential steps. For instance, Step 1 involves a Task to ``Identify the sensors available for Chiller 6 at POKMAIN site'' and specifies Agent 1 (e.g., an IoT Data Download Agent) to execute this task, with an Expected Output of ``A list of sensors available for Chiller 6 at POKMAIN site.'' Following this, Step 2 takes this output as a Dependency (\#S1) to execute the Task: ``Determine which of these sensors can detect Compressor Overheating failure,'' assigning it to a specialized Agent 2 (e.g., a FMSR Agent).

After reviewing the plan, it is translated into a dynamic, dependency-aware Workflow represented as a directed graph. This graph outlines the logical flow and potential parallel execution paths (e.g., tasks 3, 4, and 5 running concurrently) based on the sequential nature of the task dependencies. The Context-aware Execution phase then involves a specialized execution engine that manages these tasks, tracking their state, inputs (like the JSON objects and strings containing the intermediate results), and dependencies between agents. For example, the output of the first stage (ID 1) becomes a structured input for subsequent tasks (ID 2), ensuring that information is seamlessly and accurately passed between the specialized agents. The entire process culminates in a Result Summary that provides the final, actionable answer to the initial complex query. This methodology ensures traceability, modularity, and the effective integration of multiple specialized AI agents for industrial problem-solving.

\subsection{Tools used by Agents}
\label{agenttools}

In this section, we describe the development of over 15 LangChain-based tools that form the backbone of our agent framework. We follow a standardized methodology for tool construction, and, with the exception of the WO agent, all agents operate through tool-calling APIs. Table~\ref{tab:tools} lists thirteen of these tools along with their names, descriptions, and parameters. For brevity, we omit some of the lower-level parameters associated with the time-series tool suite. In case of WO agent, which is a coding agent, we needed to build a generic business driven object, as given in Table \ref{tab:business_objects}.

\begin{table*}[h!]
\centering
\caption{List of Available Tools and Their Parameters.}
\label{tab:tools}
\renewcommand{\arraystretch}{1.15}
\setlength{\tabcolsep}{3pt}
\begin{tabular}{p{2.8cm} p{5.2cm} p{5.2cm}}
\toprule
\textbf{Tool Name} & \textbf{Description} & \textbf{Parameters (Required Fields)} \\ 
\midrule

\textbf{Get Failure Modes} & 
Retrieves failure modes linked to a specific asset. & 
\texttt{asset\_name}: name of the asset. \\ 

\textbf{Get Failure Mode and Sensor Relevancy Mapping} & 
Returns relevancy mapping between failure modes and sensors for downstream tasks. & 
\texttt{input\_str}: string with asset name, failure modes, and sensors. \\ \hline

\textbf{Read Sensors From File} & 
Reads available sensors of an asset from a file and outputs sensor variable names. & 
\texttt{input\_str}: sensor file path. \\ 

\textbf{sites} & 
Retrieves a list of available sites. & 
\texttt{v\_\_args}: optional array (\textit{default: null}). \\ 

\textbf{history} & 
Returns sensor values for an asset within a given time range. & 
\texttt{site\_name}, \texttt{assetnum}, \texttt{start}, \texttt{final}. \\ 

\textbf{assets} & 
Lists all assets available at a given site. & 
\texttt{site\_name}. \\ 

\textbf{sensors} & 
Lists all sensors for an asset at a given site. & 
\texttt{site\_name}, \texttt{assetnum}. \\ 

\textbf{jsonreader} & 
Parses a JSON file and returns its content. & 
\texttt{file\_name}. \\ 

\textbf{currentdatetime} & 
Returns current date and time as JSON. & 
\texttt{v\_\_args}: optional array (\textit{default: null}). \\ \hline

\textbf{aitasks} & 
Lists available AI tasks and their methods (\texttt{task\_id}, \texttt{description}). & 
\texttt{v\_\_args}: optional array (\textit{default: null}). \\ 

\textbf{tsfmmodels} & 
Lists supported forecasting models (ID, checkpoint, description). & 
\texttt{v\_\_args}: optional array (\textit{default: null}). \\ 

\textbf{tsfm\_forecasting} & 
Forecasts sensor or KPI variables using pretrained time-series models. & 
\texttt{dataset\_path}, \texttt{model\_checkpoint}, \texttt{timestamp\_column}, \texttt{target\_columns}. \\ 

\textbf{tsfm forecasting finetune} & 
Finetunes a pretrained forecasting model on new data. & 
\texttt{dataset\_path}, \texttt{model\_checkpoint}, \texttt{timestamp\_column}, \texttt{target\_columns}. \\ 

\textbf{tsfm integrated tsad} & 
Performs time-series anomaly detection using model predictions. & 
\texttt{dataset\_path}, \texttt{timestamp\_column}, \texttt{target\_columns}. \\ 
\bottomrule
\end{tabular}
\end{table*}

\begin{table*}[ht!]
\centering
\caption{WO Agent Summary of Business Objects, Source, Role, and Number of Records}
\label{tab:business_objects}
\renewcommand{\arraystretch}{1.1} 
\begin{tabular}{p{3cm} p{2.5cm} p{6.5cm} c}
\toprule
\textbf{Business Object} & \textbf{Source} & \textbf{Role} & \textbf{Count} \\
\midrule
\multicolumn{4}{l}{\textbf{Content Objects}} \\
WorkOrder & Work Order Manager & Tracks scheduled and unscheduled maintenance tasks, categorized as preventive or corrective. & 4392 \\ \hline
Event & Aggregated by Authors & Consolidates event logs for tracking and decision-making. & 6929 \\ \hline
Alert Events & IoT Repository & Logs real-time alerts triggered by IoT sensors based on predefined conditions. & 1995 \\ \hline
Anomaly Events & ML Generated & Detects KPI deviations using machine learning for predictive maintenance. & 542 \\ 
\midrule
\multicolumn{4}{l}{\textbf{Meta/Profile Objects}} \\ 
ISO Failure Code & Developed by Authors & Standardizes failure classification for structured maintenance analysis. & 137 \\ \hline
ISO Primary Failure\_Code & Developed by Authors & Defines primary failure categories and links related secondary codes. & 68 \\ \hline
AlertRule & SME Provided & Specifies conditions for triggering alerts based on system behaviors. & 77 \\ \hline
Equipment & SME Provided & Represents industrial assets, including status and specifications. & 22 \\
\midrule
\multicolumn{4}{l}{\textbf{Relationship Causality Objects}} \\
Alert-Rule Mapping & Relationship Causality & Links alert rules to failure codes for automated diagnostics. & 46 \\ \hline
Anomaly Mapping & Relationship Causality & Associates anomalies with failure codes for predictive insights. & 12 \\
\midrule
\multicolumn{4}{l}{\textbf{Recommendation Objects}} \\
WorkOrder Recommendation & Recommendation & Suggests maintenance actions based on historical patterns. & N/A \\ \hline
\bottomrule
\end{tabular}

\vspace{4pt}
\noindent\footnotesize{\textit{Note:} The design and structure of the business objects and corresponding analysis in this section are valid for other industrial asset types, such as standby generators.}
\end{table*}

\clearpage
\newpage

\section{Benchmark Dataset Details}
\label{dbdetail}

\subsection{Datasets}\label{subsec:appendix_data}
This section zooms in on the datasets utilized by the various AssetOpsBench agents.

\textbf{Sensor Telemetry Dataset for IoT Agent and TSFM Agent}. Both IoT Agent and TSFM Agent leverage the Sensor Telemetry Dataset, which comprises sensor telemetry collected from Building Management Systems (BMS) and the SkySpark analytics platform. This dataset captures fifteen-minute interval operational data from industrial HVAC systems, specifically a fleet of chillers. Each chiller unit (e.g., Chiller 4, Chiller 14) is instrumented with a standardized suite of physical sensors that monitor key operational parameters in real-time. 

A representative subset of these sensors is summarized in Table~\ref{tab:sensors}. These sensors record various kinematic,  dynamic,  thermodynamic, electrical, and operational metrics essential to assessing the performance and health of chiller systems. Each sensor stream is accompanied by rich metadata, including sensor type, measurement units, physical location, and structured device tags that define device associations. The dataset captures realistic operational variability, encompassing noise, missing data, and seasonal patterns. As such, it provides a robust foundation for developing and benchmarking models that require temporal reasoning, fault detection, and decision-making under uncertainty.

\begin{table}[!]
\centering
\small
\caption{Representative sensors in \textbf{AssetOpsBench}.}
\label{tab:sensors}
\setlength{\tabcolsep}{4pt}
\renewcommand{\arraystretch}{1.1}
\begin{tabular}{p{0.42\columnwidth} p{0.54\columnwidth}}
\toprule
\textbf{Sensor Name} & \textbf{Description} \\
\midrule
Chiller Return Temperature & Temperature of water returning to the chiller. \\
Supply Temperature & Temperature of water exiting the chiller. \\
Power Input & Electrical power consumption. \\
Tonnage & Heat extraction rate (cooling capacity). \\
Condenser Water Supply to Chiller Temperature & Temperature of water supplied to the condenser. \\
Chiller Efficiency & Instantaneous performance metric. \\
Chiller \% Loaded & Current load as a percentage of the maximum. \\
Condenser Water Flow & Flow rate through the condenser. \\
Liquid Refrigerant Evaporator Temperature & Temperature of refrigerant in the evaporator. \\
Run Status & Binary indicator of whether the chiller is operating. \\
Setpoint Temperature & Current setpoint for chiller operation. \\
\bottomrule
\end{tabular}
\end{table}

\begin{figure*}[h]
\centering
\includegraphics[width=0.8\linewidth]{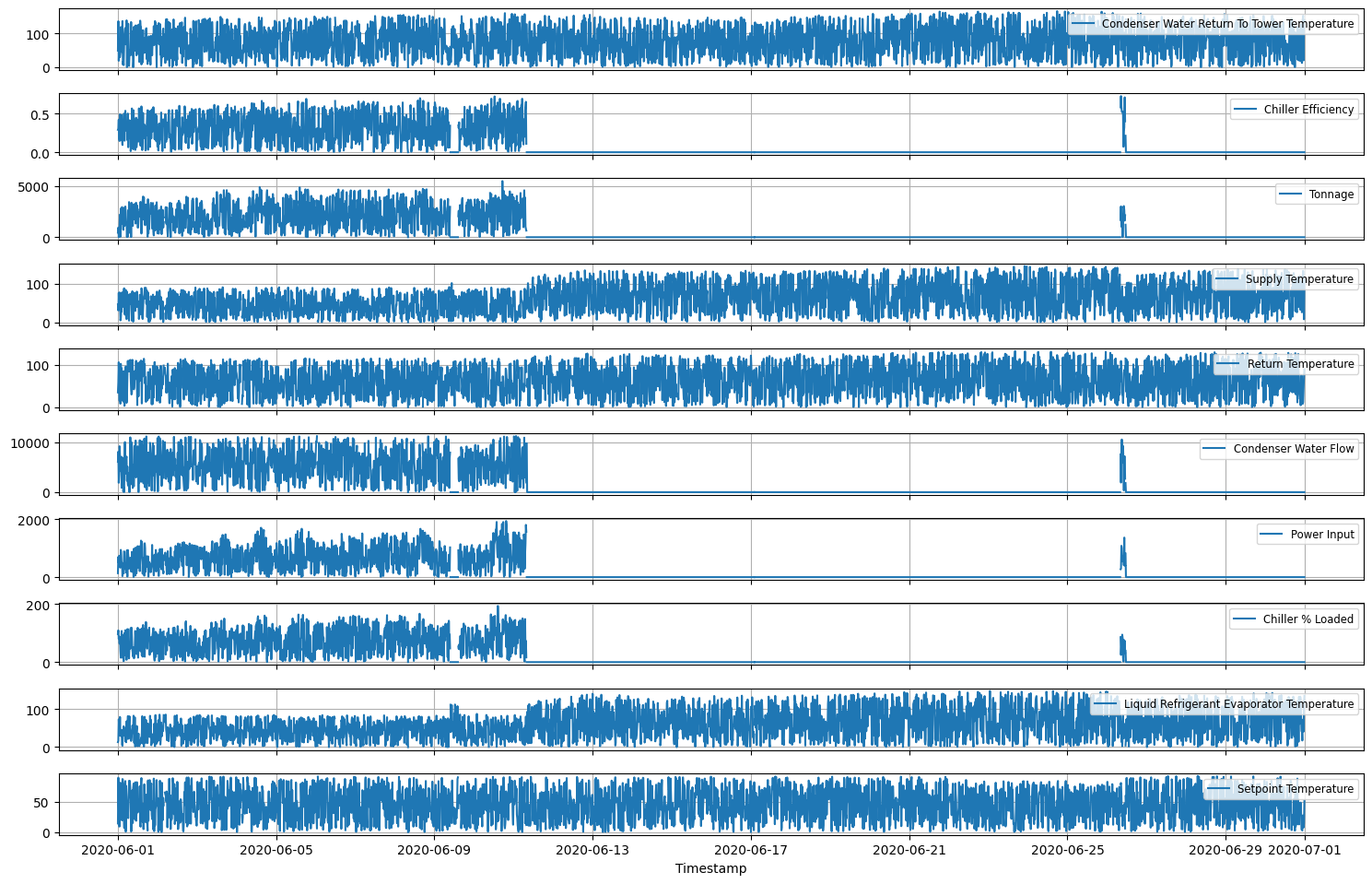}
\caption{Snapshot of time series data from Chiller 6 for June 2020. Each subplot shows an individual sensor’s trend over time.}
\label{fig:chiller6-snapshot}
\end{figure*}

As illustration, Figure~\ref{fig:chiller6-snapshot} presents layered time series subplots for key chiller sensors over a selected snapshot period in June 2020 for Chiller 6. Each subplot corresponds to one sensor variable, enabling a clear view of temporal dynamics and inter-variable behavior. This figure provides insight into the operational profile of a single chiller unit during real-world usage. 

\textbf{Failure Mode Datasets for FMSR Agent.} The failure mode datasets in AssetOpsBench are modeled using the principles of \textit{Failure Modes and Effects Analysis} (FMEA), a structured framework used in reliability engineering to identify failure risks, assess root causes and effects, and inform condition-based maintenance strategies. Each failure is defined by its mode, degradation mechanism, detection opportunity, and operational impact, enabling structured reasoning for both rule-based diagnostics and machine learning.

Failures in the dataset are annotated at the asset and subsystem levels, with a primary focus on centrifugal chillers. These failures reflect realistic degradation pathways and operational stressors derived from field experience. Each record in the failure model includes:

\begin{itemize}
    \item \textbf{Failure Location and Component:} The subsystem or part where failure occurs, such as \textit{bearings}, \textit{gearboxes}, \textit{impellers}, or \textit{lubrication systems}.
    \item \textbf{Degradation Mechanism:} The underlying physical process driving the failure, including \textit{wear}, \textit{erosion}, \textit{oil degradation}, \textit{vibration-induced fatigue}, and \textit{misalignment}.
    \item \textbf{Degradation Influences:} External or internal stressors such as \textit{run time}, \textit{two-phase process fluid}, \textit{personnel error}, or \textit{shock loading}.
    \item \textbf{Functional Failure Mode:} The resulting operational defect, such as \textit{decreased oil pressure}, \textit{audible noise}, \textit{low head pressure}, or \textit{capacity loss}.
    \item \textbf{Detection Opportunities:} Observable precursors or symptoms, including sensor readings (e.g., oil sampling, vibration signals), condition-based alarms, or inspection results.
    \item \textbf{Repair Time and Criticality:} Estimated downtime and classification of failure risk, supporting cost-based prioritization and scheduling.
    \item \textbf{Preventive Task Type:} Associated maintenance activity, such as \textit{oil analysis}, \textit{vibration analysis}, or visual inspection, tagged with effectiveness ratings and intervention intervals.
\end{itemize}

For example, \textit{bearing wear} a recurring failure across chiller subsystems may arise from lubrication failure, misalignment, or fluid shock loading. This degradation is detectable via a combination of oil analysis and vibration monitoring, with failure symptoms including increased vibration, reduced oil pressure, and audible anomalies. Similarly, impeller erosion is linked to aging and two-phase fluid exposure, typically presenting as reduced capacity and lower head pressure. More of these relations could be found in \citep{constantinides2025failuresensoriqmultichoiceqadataset}.

\textbf{Work Order Datasets for WO Agent.}  Table~\ref{tab:business_objects} provide the summary of work order datasets (as business objects) and the size for each dataset.  Those work order datasets in \textbf{AssetOpsBench} provide a structured view of maintenance activity across industrial assets, encompassing both preventive and corrective interventions using work orders.  Each work order is associated with rich contextual data including equipment metadata, failure classification codes (e.g., ISO Failure Code, ISO Primary Failure Code), event logs, sensor-triggered alerts, and machine-generated anomalies. These records are linked temporally and causally, allowing agents to reason about asset history, detect recurring failure patterns, and recommend actions based on past interventions.

\textbf{Data Quality and Operational Realism.}
The datasets in \textbf{AssetOpsBench} are derived from long-running production building management and enterprise maintenance platforms and therefore reflect the characteristics of real-world industrial data. Sensor telemetry contains missing values, flat segments, and intermittent gaps, arising primarily from genuine asset behavior such as intermittent operation and seasonal usage patterns rather than sensor faults. Maintenance work orders are event-driven and often logged retrospectively by technicians, leading to substantial variability in textual structure and imperfect temporal alignment with sensor measurements and site-specific alerts triggered from those sensors.

Rather than aggressively imputing, filtering, or normalizing these artifacts, we intentionally preserve them and treat asset inactivity and data availability as meaningful operational signals, so that model performance reflects robustness under realistic operating conditions. Downstream agents are designed to explicitly handle these characteristics. In particular, the \textbf{TSFM} agent infers operational on/off periods from sensor telemetry, aligns measurements to a common fifteen-minute grid via resampling, applies limited imputation only when appropriate, and detects anomalies using a conformal prediction framework during active operation. Similarly, the imperfect alignment between work orders, sensor data, and alerts is preserved to reflect real maintenance practice, consistent with observations from deployed work-order systems~\citep{zhou2026codereact}. Together, these design choices ensure that AssetOpsBench prioritizes operational realism over artificially curated or cleaned data.

\clearpage

\section{Scenario Creation}
\label{scedetail}
\subsection{Scenarios Creation Principles}
The scenarios in \textbf{AssetOpsBench} are designed to evaluate the capabilities required for autonomous agents operating in real industrial environments. Although grounded in real operational data and engineering practices, each scenario is intentionally framed to test a specific dimension of agent reasoning, tool interaction, and decision-making relevant to asset management. The scenarios are built around four core principles:
\begin{itemize}
\item \textbf{Reasoning and Tool Use:} Scenarios require agents to perform domain-specific reasoning such as time-based logic, schema interpretation, and multi-step tool invocation. Common failure cases include premature termination, incorrect parameter selection, or misuse of diagnostic tools.

\item \textbf{Data Handling and Forecasting:} Agents must interpret telemetry, detect anomalies, and configure appropriate models for forecasting or anomaly detection. Tasks emphasize the translation of real-world engineering intuition into ML configuration steps (e.g., model selection, training windows, thresholds).

\item \textbf{Agent Communication and Coordination:} Many scenarios simulate multi-agent workflows where the agent must ask clarifying questions, summarize findings, or coordinate subtasks. This reflects how real engineering teams collaborate during diagnostics or planning.

\item \textbf{Workflow Orchestration and Decision-Making:} Scenarios measure the agent's ability to plan complex workflows, handle dependencies, reason under uncertainty, and determine when to stop or escalate due to missing or conflicting information.
\end{itemize}

These principles ensure that scenarios remain faithful to real asset-management workflows while systematically probing the capabilities of autonomous agents.

\subsection{Scenario Examples}
We include two examples (Tables~\ref{tab:example1} and~\ref{tab:example2}) that showcase distinct behaviors of agent outputs. Although the questions are superficially similar, the required \emph{characteristic form} differs substantially.

\begin{table}[t]
\centering
\small
\setlength{\tabcolsep}{4pt}
\renewcommand{\arraystretch}{1.1}
\caption{Example knowledge query: energy prediction for Chiller 9.}
\label{tab:example1}
\begin{tabular}{@{}p{0.15\columnwidth} p{0.85\columnwidth}@{}}
\toprule
\textbf{Field} & \textbf{Description} \\
\midrule
\textbf{ID} & 507 \\
\textbf{Type} & Knowledge Query \\
\textbf{Text} & What is the predicted energy consumption for Chiller 9 in the week of 2020-04-27 based on data from the MAIN site? \\
\textbf{Characteristic form} &
\begin{minipage}[t]{0.72\columnwidth}
\vspace{2pt}
\begin{itemize}\itemsep0pt
\item Verify correct asset/location/time: \textbf{Chiller 9}, \textbf{MAIN}, week of \textbf{2020-04-27}.
\item Confirm sensor identification (power input) and historical data retrieval for the specified window.
\item Explain why forecasting cannot be performed: power input is \texttt{0.0} from 2020-04-20 to 2020-04-25 (chiller not operating), implying insufficient signal for prediction.
\end{itemize}
\vspace{2pt}
\end{minipage}
\\
\bottomrule
\end{tabular}
\end{table}

\begin{table}[t]
\centering
\small
\setlength{\tabcolsep}{4pt}
\renewcommand{\arraystretch}{1.1}
\caption{Example knowledge query: predicting energy usage for Chiller 9.}
\label{tab:example2}
\begin{tabular}{@{}p{0.15\columnwidth} p{0.85\columnwidth}@{}}
\toprule
\textbf{Field} & \textbf{Description} \\
\midrule
\textbf{ID} & 511 \\
\textbf{Type} & Knowledge Query \\
\textbf{Text} & Can you predict Chiller 9's energy usage for next week based on data from the week of 2020-04-27 at MAIN? \\
\textbf{Characteristic form} &
\begin{minipage}[t]{0.72\columnwidth}
\vspace{2pt}
\begin{itemize}\itemsep0pt
\item Verify correct asset/location and the reference week: \textbf{Chiller 9}, \textbf{MAIN}, week of \textbf{2020-04-27}.
\item Confirm sensor inventory and selection of \textit{Chiller 9 Power Input}, followed by data retrieval.
\item Report the file path where the retrieved data is stored.
\item Confirm recovery from initial analysis errors and successful fine-tuning of a \textbf{time-series forecasting} model on the retrieved data.
\item Confirm next-week predictions are generated and written to the specified output file.
\end{itemize}
\vspace{2pt}
\end{minipage}
\\
\bottomrule
\end{tabular}
\end{table}

\subsection{Scenario Comparison with Other Bench}
\label{tab:bench_comparison}
We prepare a table to compare with the literature in Table \ref{tab:benchmark_comparison}. AssetOpsBench extends prior benchmarks by incorporating temporal/dynamic queries, name disambiguation, and tool-output–driven operations. These capabilities not present in TaskBench or ITBench. Additionally, while earlier benchmarks rely on either complex tool graphs or simpler single-step tools, AssetOpsBench emphasizes multi-step tool reuse, aligning better with real industrial agent workflows.

\begin{table*}[ht]
\centering
\caption{Comparative overview of general-purpose and domain-specific benchmarks.}
\label{tab:benchmark_comparison}
\renewcommand{\arraystretch}{1.2}
\begin{tabular}{p{3.2cm} | p{3.2cm} | p{2.8cm} | p{2.8cm}}
\toprule
\textbf{Benchmark} & \textbf{TaskBench (NeurIPS~2024)} & \textbf{ITBench (ICML~2025)} & \textbf{AssetOpsBench (Ours)} \\
\midrule
Data Generation &
Tool Graph + Back-Instruct &
Manual &
Manual \\ \hline

Tool Dependency &
\checkmark &
\checkmark &
\checkmark \\ \hline

Quality Control &
LLM Self-critique + Rule-based &
Human Verification &
Human Verification \\ \hline

Evaluation &
Task Decomposition + Tool Selection + Parameter Prediction &
ReActive Planning + Tool Selection &
ReActive Planning + Tool Selection + Parameter Prediction \\ \hline

Tool Complexity &
Single tool to complex tool graph &
-- &
Multiple tools; same tools can be called multiple times \\ \hline

Dataset Scale &
17,331 samples &
141 scenarios &
141 scenarios \\ \hline

Temporal / Dynamic Query &
$\times$ &
$\times$ &
\checkmark \\ \hline

Name Disambiguation &
$\times$ &
$\times$ &
\checkmark \\ \hline

Tools Output Operation &
$\times$ &
$\times$ &
\checkmark \\ \hline
\bottomrule
\end{tabular}
\end{table*}

\subsection{User Study Realism Analysis}
\label{sec:reliability}
\begin{figure}[h]
    \centering
    \includegraphics[width=0.95\linewidth]{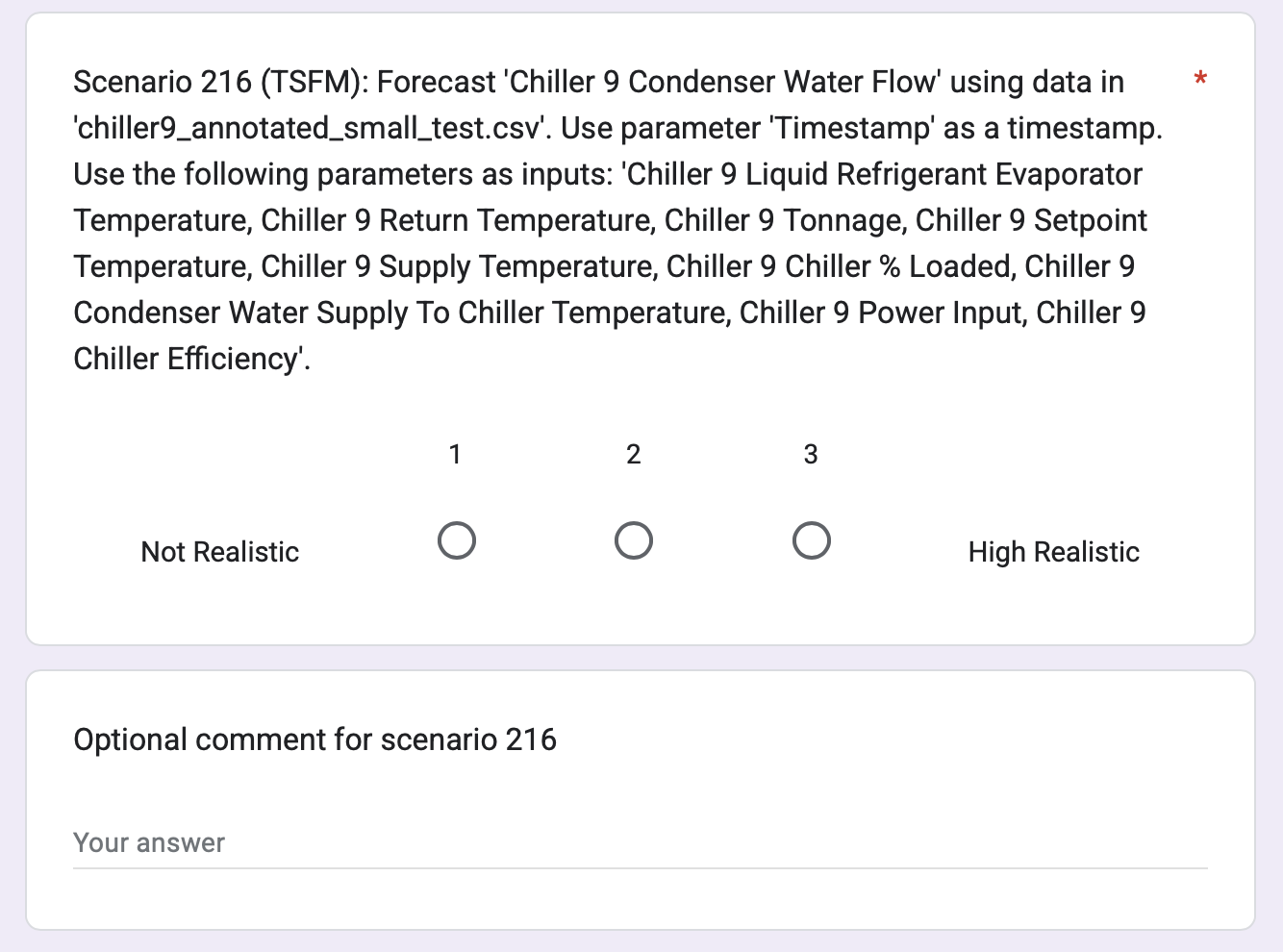}
    \caption{Representative example of Scenario for Collecting user feedback}
    \label{fig:realisum}
\end{figure}

To quantitatively assess the realism of AssetOpsBench scenarios, we conducted a human evaluation study. We randomly selected 25 representative scenarios covering four categories: IoT queries, time-series forecasting (TSFM), work orders/events, and failure mode reasoning (FMSR). Participants were domain experts, including reliability engineers, maintenance engineers, and data scientists familiar with condition-based monitoring and predictive maintenance. Each participant evaluated scenarios using a 3-point scale (\textit{1 = Not Realistic, 2 = Realistic, 3 = High Realistic}) and could optionally provide qualitative comments. Background questions captured participants' role, years of experience, and familiarity with predictive maintenance. Responses were collected via a Google Form as shown in Figure \ref{fig:realisum}. 

The following metrics were computed to assess internal consistency, inter-rater agreement, and the reliability of aggregated scores.

\subsubsection{Reliability Metrics}

The internal consistency of the 25 scenario ratings is excellent, as indicated by a Cronbach's alpha of 0.887. The ICC(1) value of 0.233 reflects moderate agreement at the individual-participant level, whereas ICC(2) of 0.882 demonstrates that the aggregated ratings across participants are highly reliable. Fleiss' kappa of 0.209 indicates slight-to-fair categorical agreement among participants, which is consistent with the subjective nature of realism judgments. Overall, while individual participants may vary in their ratings, the averaged scores per scenario provide a stable and trustworthy measure of perceived realism.

\section{LLM-As-Judge Evaluation}

\subsection{System Prompt}\label{llm-as-judge-prompt}
Table \ref{tab:questionnaire} is the prompt instruction to the evaluation agent, which outlines the specific evaluation dimensions, constraints, and response formatting
guidelines that the model follows when scoring task outputs. The evaluation criteria is also provided to human judges which ensures consistency across evaluations.

\begin{table*}[t]
\centering
\small
\begin{tabular}{|>{\columncolor{blue!10}}p{13cm}|}
\hline
You are a critical reviewer tasked with evaluating the effectiveness and accuracy of an AI agent's response to a given task. Your goal is to determine whether the agent has successfully accomplished the task correctly based on the expected or characteristic behavior.

\textbf{Evaluation Criteria:}

1. \textbf{Task Completion}:
\newline - Verify whether the agent executed all required actions (e.g., using the correct tools, retrieving data, performing the necessary analysis).
\newline - Ensure the response aligns with the predefined expected behavior for task completion.

2. \textbf{Data Retrieval \& Accuracy}:
\newline - Confirm that the correct asset, location, time period, and sensor (if applicable) were used.
\newline - Check that the retrieved data and results (forecasting, anomaly detection, etc.) are correct and consistent with the task requirements.

3. \textbf{Generalized Result Verification}:
\newline - Task Type Verification: Assess if the agent returned the expected results for the task type (forecasting, anomaly detection, classification, etc.).
\newline - Forecasting: Ensure forecasts cover the specified future period.
\newline - Anomaly Detection: Verify that anomalies were correctly detected when expected.
\newline - Other Tasks (e.g., classification): Check that results match expected format and values.
\newline - Comparison with Expected Output: Validate that results match the characteristic answer.
\newline - Data Integrity: Ensure correct data (sensor, time period) was used and output format is consistent.

\textbf{Inputs:}
\newline Question: \textcolor{red}{\{question\}}
\newline Characteristic Answer (Expected Behavior): \textcolor{red}{\{characteristic\_answer\}}
\newline Agent's Thinking: \textcolor{red}{\{agent\_think\}}
\newline Agent's Final Response: \textcolor{red}{\{agent\_response\}}

\textbf{Output Format:}
\newline Provide your review strictly in JSON format without any additional text or Markdown.
\newline \{
\newline \quad "task\_completion": true/false,
\newline \quad "data\_retrieval\_accuracy": true/false,
\newline \quad "generalized\_result\_verification": true/false,
\newline \quad "suggestions": "Optional. Recommended actions to improve the agent's response if needed."
\newline \}
\newline (END OF RESPONSE)

Evaluate the agent's performance according to the above criteria.
\\
\hline
\end{tabular}
\caption{Prompt instruction for LLM-as-a-judge
evaluation agent}
\label{tab:questionnaire}
\end{table*}

\subsection{Human Validation}

\begin{figure}
    \centering
    \includegraphics[width=0.7\linewidth]{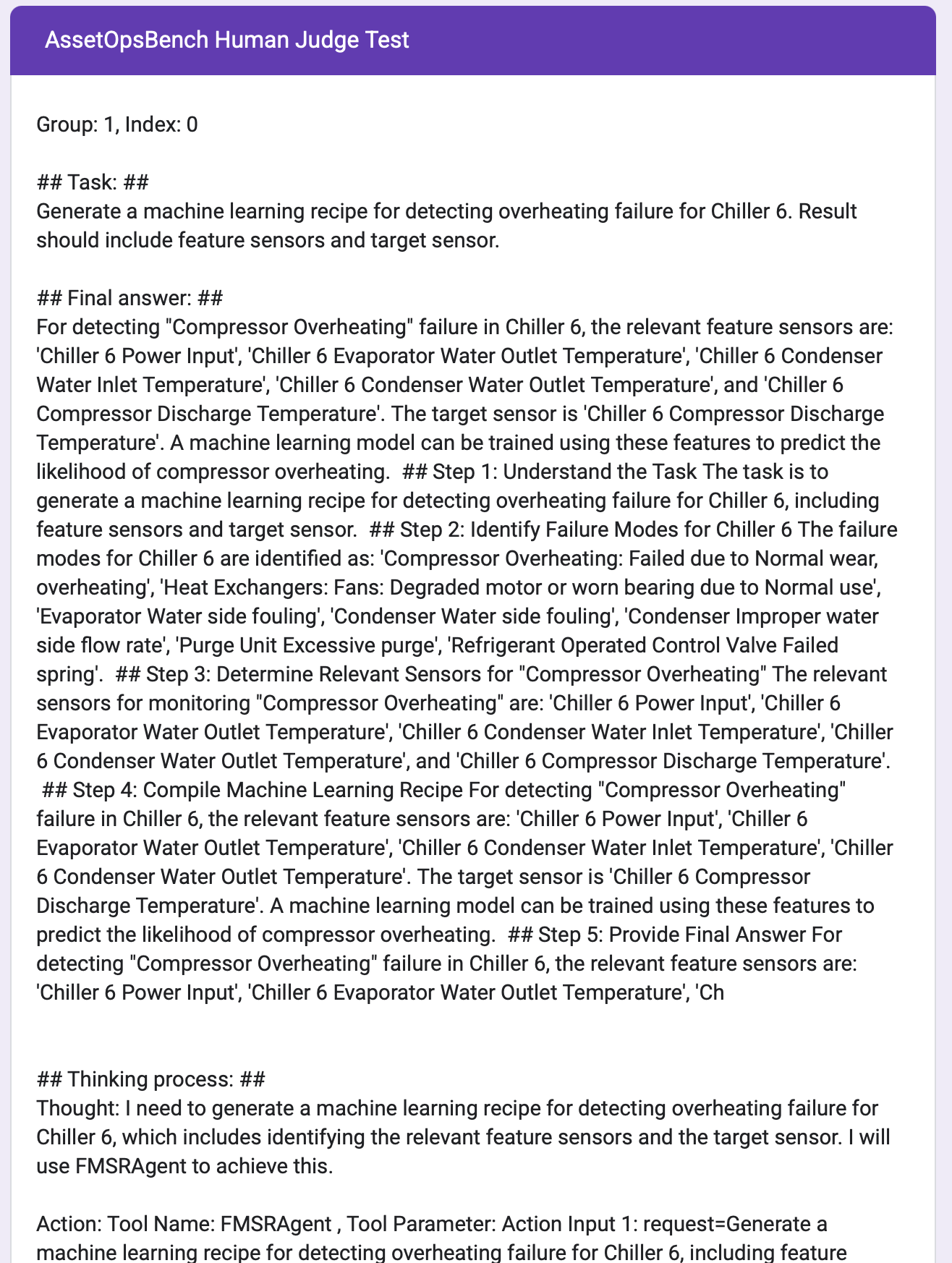}
    \caption{Google Forms: questionnaire to domain experts for human validation}
    \label{fig:googleform}
\end{figure}

We conducted human validation of LLM-as-Judge using Google Forms. As shown in Figure \ref{fig:googleform}, domain experts were provided with the original task description, the agent’s reasoning and final answer, and a checklist spanning six evaluation dimensions. Each dimension was assessed using a binary (True/False) judgment, enabling structured comparison between human and model evaluations. We distributed four forms, each containing 10 samples, resulting in 240 total data points. The aggregated results are reported in Section \ref{leaderboard}.

\section{Reference-Based Evaluation}
\subsection{Ground Truth Preparation}
To ensure that each scenario can be objectively evaluated, we first construct a ground-truth specification that precisely defines the expected reasoning steps and final answer. The example in Figure \ref{fig:fmsr_example} illustrates the task of the FMSR agent in which the system must retrieve sensor names associated with a wind turbine. The ground truth includes the task description, the required planning step, and the exact sequence of execution actions that a correct agent should follow. By explicitly defining the operations — such as calling get available sensor information with the asset name “Wind Turbine”, the
ground truth provides a verifiable reference trace.
\begin{figure*}[t]
\centering
\small
\fbox{%
\begin{minipage}{0.8\textwidth}
\ttfamily
\raggedright
\{\\
\ \ "id": 105,\\
\ \ "type": "FMSR",\\
\ \ "deterministic": false,\\
\ \ "characteristic\_form": "the answer should contain a list of sensor names for asset wind turbine.",\\
\ \ "text": "Provide some sensors of asset Wind Turbine.",\\
\ \ "planning\_steps": ["Provide some sensors of asset Wind Turbine."],\\
\ \ "execution\_steps": [\\
\ \ \ \ \{"name": "get\_available\_sensor\_information", "action": "Get Available Sensor Information",\\
\ \ \ \ \ "arguments": "Wind Turbine", "outputs": "[a list of sensor names]"\},\\
\ \ \ \ \{"name": "finish", "action": "Finish", "arguments": "", "outputs": ""\}\\
\ \ ],\\
\ \ "execution\_links": [\{"source": "get\_available\_sensor\_information", "target": "finish"\}]\\
\}\\
\end{minipage}}
\caption{An Example FMSR task specification and ground truth.}
\label{fig:fmsr_example}
\end{figure*}

\subsection{ROUGE Scoring Specification}
ROUGE metrics used include:  
\begin{itemize}
    \item \texttt{rouge1}: unigram (1-gram) overlap between generated and reference outputs.
    \item \texttt{rouge2}: bigram (2-gram) overlap.
    \item \texttt{rougeL}: longest common subsequence between generated and reference sequences.
    \item \texttt{rougeLsum}: line-wise longest common subsequence for multi-line outputs.
\end{itemize}

\section{Additional Experimental Results}

\subsection{Planning Capability Analysis}

We extended our evaluation to include models that were not part of the original benchmark, specifically Anthropic Claude variants (\texttt{claude-3-7-sonnet}, \texttt{claude-4-sonnet}) and GCP Gemini (\texttt{\seqsplit{gemini-2.5-pro}}). Model performance was evaluated using planning accuracy metrics (BERTScore, ROUGE, and alignment with ground-truth plans), consistent with our original execution-accuracy leaderboard. We also analyzed planned step statistics to understand model behavior in generating task plans.

For each model, we compute a \textit{combined planning score} $S_\text{combined}$ that integrates BERTScore ($B$) and ROUGE-L ($R$):
\[
S_\text{combined} = \frac{B + R}{2},
\]
where $B \in [0,1]$ is the average BERTScore between the model-generated plan and the ground-truth plan, and $R \in [0,1]$ is the average ROUGE-L F1 score. This provides a balanced measure of both semantic similarity (via BERTScore) and sequence-level overlap (via ROUGE-L).

We summarize planning quality and plan length statistics across models in Tables~\ref{tab:planning_accuracy} and~\ref{tab:planned_steps}, respectively. The following are a few key insights from the results.

\begin{table}[t]
\centering
\setlength{\tabcolsep}{4.5pt}
\renewcommand{\arraystretch}{1.08}
\caption{Planning accuracy (Avg $\pm$ Std) for all evaluated models.}
\label{tab:planning_accuracy}
\begin{tabular}{@{}lcc@{}}
\toprule
\textbf{Model} & \textbf{Avg $\pm$ Std} & \textbf{\#Q} \\
\midrule
mistral-medium-2505 & 0.620 $\pm$ 0.063 & 141 \\
gemini-2.5-pro & 0.615 $\pm$ 0.068 & 141 \\
gpt-oss-120b & 0.606 $\pm$ 0.077 & 141 \\
mistral-small-3-1-24b & 0.604 $\pm$ 0.062 & 141 \\
claude-4-sonnet & 0.595 $\pm$ 0.068 & 141 \\
llama-3-405b-instruct & 0.588 $\pm$ 0.074 & 141 \\
claude-3-7-sonnet & 0.571 $\pm$ 0.071 & 141 \\
llama-4-maverick-17b & 0.558 $\pm$ 0.071 & 141 \\
gpt-5-2025-08-07 & 0.544 $\pm$ 0.092 & 141 \\
granite-3-3-8b-instruct & 0.529 $\pm$ 0.067 & 141 \\
llama-3-3-70b-instruct & 0.522 $\pm$ 0.068 & 141 \\
\bottomrule
\end{tabular}
\end{table}

\begin{table}[t]
\centering
\small
\setlength{\tabcolsep}{3.5pt}
\renewcommand{\arraystretch}{1.06}
\caption{Planned step statistics for all evaluated models.}
\label{tab:planned_steps}
\resizebox{\columnwidth}{!}{%
\begin{tabular}{@{}lccccc@{}}
\toprule
\textbf{Model} & \textbf{Avg steps $\pm$ Std} & \textbf{Min} & \textbf{Max} & \textbf{Zero} \\
\midrule
llama-3-405b-instruct & 3.14 $\pm$ 1.84 & 1 & 9 & 0 \\
llama-3-3-70b-instruct & 6.55 $\pm$ 1.55 & 3 & 12 & 0 \\
llama-4-maverick-17b & 4.34 $\pm$ 1.80 & 1 & 9 & 0 \\
granite-3-3-8b-instruct & 5.56 $\pm$ 2.44 & 2 & 30 & 0 \\
gpt-oss-120b & 1.91 $\pm$ 1.21 & 1 & 10 & 0 \\
mistral-medium-2505 & 2.38 $\pm$ 1.04 & 1 & 5 & 0 \\
mistral-small-3-1-24b & 2.77 $\pm$ 1.33 & 1 & 6 & 0 \\
claude-3-7-sonnet & 3.10 $\pm$ 1.15 & 1 & 5 & 0 \\
gpt-5-2025-08-07 & 2.33 $\pm$ 1.16 & 0 & 5 & 1 \\
gemini-2.5-pro & 1.87 $\pm$ 1.01 & 1 & 5 & 0 \\
claude-4-sonnet & 2.45 $\pm$ 1.34 & 1 & 5 & 0 \\
\bottomrule
\end{tabular}}
\end{table}

\begin{itemize}
    \item \textbf{Top-performing models:} \texttt{mistral-medium-2505} achieves the highest planning accuracy (0.620 ± 0.063) and produces concise, low-variance plans (2.38 ± 1.04 steps), combining high performance with efficiency. \texttt{gemini-2.5-pro} is also highly competitive (0.615 ± 0.068).
    \item \textbf{Anthropic Claude models:} Both \texttt{claude-3-7-sonnet} and \texttt{claude-4-sonnet} show solid planning accuracy with moderate plan lengths (~2.5–3 steps) and low variance, indicating reliable reasoning and execution alignment.
    \item \textbf{Instruction tuning matters:} Medium-sized instruction-tuned models (\texttt{mistral-medium}, \texttt{mistral-small}) consistently produce efficient plans with low variance, outperforming larger models with longer, more variable plans (\texttt{llama-3-3-70b}, \texttt{granite-3-3-8b}).
    \item \textbf{Step length vs. performance:} Shorter plans with low variance generally correlate with higher planning accuracy, while overly long plans may introduce redundancy without improving alignment with ground-truth executions.
    \item \textbf{Consistency vs. variability:} High-variance models (\texttt{\seqsplit{gpt-5-2025-08-07}}, \texttt{granite-3-3-8b}) occasionally generate very long or empty plans, which may reduce reliability despite moderate average scores.
\end{itemize}

\textbf{Conclusion:} The unified analysis demonstrates that medium-sized, instruction-tuned models offer the best balance of planning accuracy and step efficiency, while the inclusion of Claude and Gemini models extends benchmark coverage and validates performance trends. These results are consistent with our original execution-accuracy leaderboard, confirming the robustness of the benchmark for evaluating reasoning-capable language models.

\subsection{Cost Analysis}
\label{sec:cost_analysis}

Table~\ref{tab:cost} compares the estimated cost of running the full 140+ utterance task suite under the \textbf{Agent-as-Tool} paradigm. We report the average tokens consumed per task and the resulting total cost for each provider/model configuration.

\begin{center}
\small
\captionof{table}{Runtime and estimated cost for executing 140+ utterance tasks using the Agent-as-Tool paradigm.}
\label{tab:cost}
\setlength{\tabcolsep}{6pt}
\renewcommand{\arraystretch}{1.15}
\begin{tabular}{@{}l l r r@{}}
\toprule
\textbf{LLM} & \textbf{Provider} & \textbf{Avg.\ tokens} & \textbf{Total cost (USD)} \\
\midrule
\texttt{gpt-4.1} & OpenAI & 3{,}664 & \$300 \\
\texttt{llama-4-maverick} & watsonx & 3{,}730 & \$130 \\
\bottomrule
\end{tabular}
\end{center}

As shown in Table~\ref{tab:cost}, token usage is comparable across the two models, but the total estimated cost differs substantially, reflecting provider pricing rather than task-specific token demand. In our setup, \texttt{llama-4-maverick} yields a lower total cost while maintaining similar average tokens per task, which can be advantageous when scaling evaluation runs or performing repeated ablations.

\subsection{Uncertainty Analysis}
\label{uana}

As discussed in Section~\ref{sec:Setting}, we run the evaluation agent five times to estimate the stability of rubric-based metrics. Table~\ref{tab:interrateragreement} reports inter-rater agreement across the five runs, together with the derived uncertainty, computed as $1-\text{agreement}$.

\begin{center}
\captionof{table}{Inter-rater agreement and derived uncertainty across five evaluation runs.}
\label{tab:interrateragreement}
\setlength{\tabcolsep}{6pt}
\renewcommand{\arraystretch}{1.15}
\begin{tabular}{@{}lcc@{}}
\toprule
\textbf{Metric} & \textbf{Agreement} & \textbf{Uncertainty} \\
\midrule
Task Completion & 0.9731 & 2.69\% \\
Data Retrieval Accuracy & 0.9697 & 3.03\% \\
Generalized Result Verification & 0.9681 & 3.19\% \\
\bottomrule
\end{tabular}
\end{center}

Overall, agreement is consistently high ($\geq$0.968), corresponding to uncertainties below 3.2\%, which indicates that the LLM-as-judge evaluation is stable under repeated runs.

\section{Ablation Studies}
\label{astudy}
\subsection{Distractor Agents}

We introduce 10 distractor agents to intentionally increase the complexity and ambiguity faced by a global agent. Table~\ref{tab:agent_type_roles} summarizes these distractors by domain and role, spanning general-purpose behaviors (e.g., echoing inputs, off-topic responses) and domain-specific capabilities (e.g., predictive maintenance, sensor summarization, edge ML deployment). This taxonomy increases the realism of multi-agent environments by supporting modular integration while injecting controlled confusion.

\begin{table*}[t]
\centering
\small
\renewcommand{\arraystretch}{1.08}
\caption{Distractor agent types and their roles.}
\label{tab:agent_type_roles}

\begin{tabularx}{0.95\textwidth}{@{}l l X@{}}
\toprule
\textbf{Agent} & \textbf{Domain} & \textbf{Role} \\
\midrule
\texttt{EchoAgent} & General & Repeats the input verbatim (debugging / coherence checks). \\
\texttt{OffTopicAgent} & General & Returns unrelated facts when a query is off-topic or unrecognized. \\
\texttt{CustomerSupportAgent} & Support Ops & Handles password resets, login issues, and service availability. \\
\texttt{SREAgent} & Site Reliability & Diagnoses performance degradation, downtime, and infrastructure issues. \\
\texttt{FrontendDevAgent} & Software Eng. & Assists with UI/UX, React, and rendering bugs. \\
\texttt{HRPolicyAgent} & Human Resources & Answers policy questions (leave, benefits, compliance). \\
\texttt{SensorDataSummarizer} & Industrial IoT & Summarizes sensor time series (trends, anomalies). \\
\texttt{HistoricalTrendsAgent} & Analytics & Extracts historical patterns for failure analysis/optimization. \\
\texttt{EdgeMLAgent} & Edge Computing & Recommends strategies for ML deployment on constrained hardware. \\
\texttt{RULPredictorAgent} & Predictive Maint. & Estimates remaining useful life from sensor and degradation signals. \\
\bottomrule
\end{tabularx}

\end{table*}
Across 99 scenarios, we compare model performance with and without distractor agents to evaluate robustness under noisy agent-selection spaces. Table~\ref{tab:model_distractor_comparison} reports rubric-based scores. \texttt{\seqsplit{gpt-4.1-2025-04-14}} remains the strongest overall, with only a small drop under distractors (e.g., task completion $52\rightarrow48$). Among open-weight models, \texttt{llama-4-maverick} is the most robust and even improves with distractors (task completion $46\rightarrow48$, verification $46\rightarrow49$), whereas \texttt{mistral-large} and \texttt{llama-4-scout} degrade more noticeably. Interestingly, \texttt{llama-3-405b-instruct} improves across all three metrics, suggesting strong corrective behavior under ambiguity, while \texttt{granite-3-3-8b-instruct} remains stable but at a lower accuracy level.

\begin{table*}[t]
\centering
\small
\setlength{\tabcolsep}{7pt}
\renewcommand{\arraystretch}{1.08}
\caption{Model performance with vs.\ without distractor agents (99 scenarios).}
\label{tab:model_distractor_comparison}
\begin{tabular}{@{}l l c c c@{}}
\toprule
\textbf{Model} & \textbf{Setting} & \textbf{Task Completion} & \textbf{Data Accuracy} & \textbf{Result Verification} \\
\midrule
\texttt{gpt-4.1-2025-04-14} & Without distractors & 52 & 57 & 55 \\
                           & With distractors    & 48 & 56 & 54 \\
\midrule
\texttt{granite-3-3-8b-instruct} & Without distractors & 40 & 44 & 41 \\
                                & With distractors    & 40 & 44 & 41 \\
\midrule
\texttt{mistral-large} & Without distractors & 42 & 46 & 43 \\
                       & With distractors    & 40 & 44 & 41 \\
\midrule
\texttt{llama-3-405b-instruct} & Without distractors & 41 & 41 & 38 \\
                              & With distractors    & 44 & 44 & 44 \\
\midrule
\texttt{llama-3-3-70b-instruct} & Without distractors & 38 & 43 & 34 \\
                               & With distractors    & 41 & 43 & 36 \\
\midrule
\texttt{llama-4-maverick} & Without distractors & 46 & 49 & 46 \\
                          & With distractors    & 48 & 49 & 49 \\
\midrule
\texttt{llama-4-scout} & Without distractors & 45 & 44 & 46 \\
                       & With distractors    & 40 & 40 & 40 \\
\bottomrule
\end{tabular}

\vspace{2pt}
\footnotesize\emph{Note:} Higher is better. Scores are reported as percentages. 
\end{table*}

\subsection{Impact of in-context examples}
Table \ref{tab:incontext_comparison} compares \texttt{gpt-4.1} and \texttt{granite-3-3-8b} with and without in-context examples on a subset of single-agent benchmark tasks ($n=65$). Consistent with our main findings, in-context examples are critical for enabling effective ReAct-style reasoning and coordination.

Removing in-context examples causes a large performance drop for both models: \texttt{gpt-4.1} decreases from an average of 80\% (with context) to 33\% (without), while \texttt{granite-3-3-8b} falls from 60\% to 3\%. These results reinforce that in-context examples are essential for ReAct-style reasoning in LLM-based agents. We do not include work-order and end-to-end tasks here because their baseline performance is already low.

\subsection{Baseline using Single-Agent}
Instead of using four domain-specific sub-agents and an orchestration agent, we build a tool-calling ReAct agent with a single prompt as a baseline, giving it access to tools and in-context examples from all agents. In doing so, we increase the complexity of the problem, as it must handle many tools as well as an expanded context. We run a default LLM, \texttt{llama-4-maverick}, on all 141 scenarios. As a single-agent baseline, it achieves task completion of 26.95\%, data retrieval accuracy of 34.04\%, and generalized result verification of 28.37\%. Under the Agent-As-Tool setup, the same model achieves roughly two-fold improvements (See Figure \ref{fig:agent-as-tool}).

\FloatBarrier        
\begin{table*}[!ht]
\centering
\small
\setlength{\tabcolsep}{5pt}
\renewcommand{\arraystretch}{1.12}
\caption{Impact of in-context examples on single-agent tasks ($n=65$). Scores are percentages.}
\label{tab:incontext_comparison}
\begin{tabular}{@{}l c c c c@{}}
\toprule
\textbf{Model} & \textbf{Examples} & \textbf{Completion} & \textbf{Retrieval} & \textbf{Verification} \\
\midrule
\texttt{gpt-4.1} & Yes & 52 & 57 & 55 \\
                 & No  & 22 & 21 & 24 \\
\addlinespace[2pt]
\texttt{granite-3-3-8b} & Yes & 40 & 44 & 41 \\
                        & No  & 2  & 3  & 3 \\
\bottomrule
\end{tabular}
\end{table*}

\end{document}